\pdfoutput=1

\documentclass[11pt,x11names,table]{article}
\usepackage{setspace}
\usepackage{misc/emnlp2023}

\usepackage{times}
\usepackage{xcolor}
\usepackage{graphicx}
\usepackage[T1]{fontenc}
\usepackage[utf8]{inputenc}
\usepackage{microtype}
\usepackage{booktabs}
\usepackage{todonotes}
\usepackage{amsmath,amssymb}
\usepackage{cleveref}
\usepackage{xspace}
\usepackage{soul} %
\usepackage{caption}
\usepackage{multirow}
\usepackage{ulem}
\usepackage{float}
\usepackage{array}
\usepackage{bm}
\usepackage{xfrac}
\usepackage{tcolorbox} %
\usepackage[shortlabels]{enumitem}
\usepackage[outline]{contour}%
\usepackage{tikz}
\usepackage{algorithm,algpseudocode,algorithmicx}

\crefname{algorithmB}{algorithm}{algorithms}
\Crefname{algorithmB}{Algorithm}{Algorithms}
\DeclareCaptionType{algorithmB}[Algorithm][List of algorithms]

\setlist[itemize]{noitemsep,left=0mm}

\usepackage{tabularx}

\crefname{lstlisting}{listing}{listings}
\Crefname{lstlisting}{Listing}{Listings}
\crefname{equ}{equation}{equations}
\Crefname{equ}{Equation}{Equations}
\Crefname{algorithm}{Algorithm}{Algorithms}
\crefname{example}{example}{examples}
\Crefname{example}{Example}{Examples}
\crefname{prompt}{prompt}{prompts}
\Crefname{prompt}{Prompt}{Prompts}
\DeclareCaptionType{example}[Example][List of examples]
\DeclareCaptionType{prompt}[Prompt][List of prompts]
\DeclareCaptionType{equ}[Equation][List of equations]

\DeclareMathOperator*{\argmax}{arg\,max}

\usepackage{marginnote}

\definecolor{TodoColor}{rgb}{1,0.7,0.6}
\definecolor{TodoColor2}{rgb}{0.7,0.7,0.9}
\definecolor{TodoColor3}{rgb}{0.5,0.8,0.5}

\algrenewcommand\algorithmicindent{1.0em}

\makeatletter\def\Hy@Warning#1{}\makeatother
\let\svthefootnote\thefootnote
\newcommand\blankfootnote[1]{%
  \let\thefootnote\relax\footnotetext{#1}%
  \let\thefootnote\svthefootnote%
}

\graphicspath{{img/}}

\title{
    Early-Exit and Instant Confidence Translation Quality Estimation
}

\author{
Vilém Zouhar$^1$ \quad
Maike Züfle$^2$ \quad
Beni Egressy$^3$ \\
\bf Julius Cheng$^4$ \quad
Mrinmaya Sachan$^1$ \quad
Jan Niehues$^2$ \\[0.5em]
$^1$ETH Zurich \quad
$^2$Karlsruhe Institute of Technology \\
$^3$Heidelberg Institute for Theoretical Studies \quad 
$^4$University of Cambridge
}

\begin{document}

\maketitle

\maketitle

\begin{abstract}

Quality estimation is omnipresent in machine translation, for both evaluation and generation.
Unfortunately, quality estimation models are often opaque and computationally expensive, making them impractical to be part of large-scale pipelines.
In this work, we tackle two connected challenges:
(1) reducing the cost of quality estimation at scale, and (2) developing an inexpensive uncertainty estimation method for quality estimation.
To address the latter, we introduce \textit{Instant Confidence COMET}, an uncertainty-aware quality estimation model that matches the performance of previous
approaches at a fraction of their costs.
We extend this to \textit{Early-Exit COMET}, a quality estimation model that can compute quality scores and associated confidences already at early model layers, allowing us to early-exit computations and reduce evaluation costs.
We also apply our model to machine translation reranking.
We combine \textit{Early-Exit COMET} with an upper confidence bound bandit algorithm to find the best candidate from a large pool without having to run the full evaluation model on all candidates.
In both cases (evaluation and reranking) our methods reduce the required compute by 50\% with very little degradation in performance.
Finally, we show how
\textit{Instant Confidence COMET} can be used to decide which translations a human evaluator should score rather than relying on the COMET score.
\end{abstract}

\section{Introduction}
\blankfootnote{\hspace{-1em}$^\dagger$Code and models: \href{https://github.com/zouharvi/COMET-early-exit}{github.com/zouharvi/COMET-early-exit}
}

Machine Translation (MT) has made significant progress, with state-of-the-art models achieving increasingly high-quality translations \citep{kocmi-etal-2023-findings,kocmi-etal-2024-findings}.
However, as reliance on automatic translation grows, so does the need for robust evaluation metrics.
Quality estimation is one such approach, allowing for the automated assessment of translations without the need for reference texts \citep{tamchyna-2021-deploying,freitag-etal-2023-results,freitag-etal-2024-llms}.
Recently, quality estimation models have also been finding their way into the decoding process \citep{10.1162/tacl_a_00491, finkelstein2024mbrqefinetuningtrainingtime}, in particular for reranking translation candidates \citep{shen-etal-2004-discriminative,freitag-etal-2022-high,cheng2024bayesianoptimizationapproachmachine}.

\begin{figure}[t]
    \centering
    \includegraphics[width=0.9\linewidth]{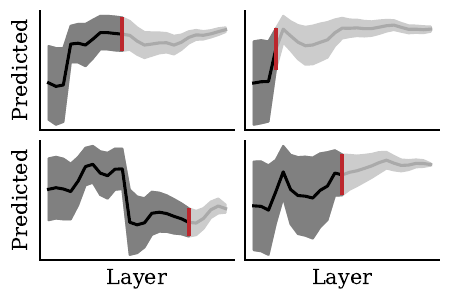}

    \vspace{-3mm}
    \caption{Progression of predicted quality estimation score (dark line) and instant confidence estimation (shaded area) along the quality estimation model computation for four examples from the test set. Layer corresponds to compute cost.
    Red line \textcolor{red!80!black}{\bf |} stops computation because the confidence is high enough (early-exit).
    }
    \label{fig:14-plot_conf_individual}
\end{figure}

Despite its promise, we identify two challenges that prevent the widespread adoption of existing quality estimation methods.
First, current methods typically provide a single best-guess quality score, which may be inadequate when uncertainty is high.
In critical industry applications, such as legal, medical, or diplomatic contexts, misjudging translation quality can have dire consequences and high scores with very low confidence should better be ignored.
Knowing the certainty of a prediction is important for decision-making, such as deferrals or routing \citep{zhang2025leveraginguncertaintyestimationefficient,farinhas2025translatesmarthardcascaded} or
for efficient human evaluation, reserving only uncertain examples for human judgment.
Previous work has used techniques such as Monte Carlo dropout \citep{glushkova-etal-2021-uncertainty-aware} to estimate confidence intervals for quality scores.
Although promising, these methods require the model to be run multiple times for each input to generate different stochastic predictions, leading to a substantial computational burden. This limits their practical use in large-scale systems.
On the other hand, direct confidence estimation \citep{zerva-etal-2022-disentangling} requires separate models for the quality score and the confidence score, which carries a factor two overhead.

Second, the increasing size and complexity of quality estimation models makes it computationally expensive to evaluate hundreds of candidates at once \citep{guerreiro-etal-2024-xcomet}, severely limiting the scalability.
This makes it difficult to use quality estimation in time-sensitive scenarios, where fast evaluations are critical.
Additionally, the high computational costs restrict the deployment in resource-constrained environments, hindering practical use in large-scale machine translation.

To overcome the first challenge, we propose \textit{Instant Confidence COMET}, a model that jointly predicts translation quality and an associated uncertainty, without increasing the computational cost.
We demonstrate in \Cref{sec:goal_instant_confidence} that the model can produce accurate confidence scores that strongly correlate with true prediction errors, without affecting the quality estimation performance.

We also use the confidence mechanism for our new \textit{Early-Exit COMET}, a model that estimates translation quality already in the earlier layers.
Examples of how score and confidence predictions develop throughout the layers are shown in \Cref{fig:14-plot_conf_individual}.
Early-Exit COMET uses the confidence score to determine whether an early prediction suffices or additional evaluation layers are needed.
In \Cref{sec:goal_earlyexit1} we combine it with a simple threshold-based early-exit algorithm for faster sample-level quality estimation.
Finally, in \Cref{sec:goal_earlyexit2} we use Early-Exit COMET for machine translation reranking.
In this task, the machine translation model generates many candidates and the goal is to find the best one with as little compute as possible.
We combine Early-Exit COMET with an upper confidence bound bandit algorithm for this reranking process and show that it outperforms strong random and logprob-based baselines.

Another use for confidence scores by \textit{Instant Confidence COMET} is to decide whether a particular translation needs to be scored by humans or if the automated metric suffices.
We show positive results for this mixed human-metric annotation in \Cref{sec:deferral}.

In addition to our results and code, we release the pre-trained Early-Exit and Instant Confidence quality estimation models publicly.

\section{Instant Confidence COMET}
\label{sec:goal_instant_confidence}
When a quality estimation model produces an output, such as 85, it is not clear whether this is just the model's best guess or truly an accurate assessment of the quality.
Having this additional information is important for decision making \citep{zhang2025leveraginguncertaintyestimationefficient,farinhas2025translatesmarthardcascaded}, such as in commercial translation pipelines where high-scoring translations are marked as requiring very little human attention. If the high score is very uncertain, then it would be safer to have a human double-check these translations.
Similarly, when designing budget-efficient human evaluation pipelines \citep{zouhar2025selectdatapointsefficienthuman}, human labor should be directed to examples where automated metrics are the least confident.

\paragraph{Preliminaries.}
A quality estimation model is a function $f$ that given a source $s$ and a translation $t$, computes an estimate of the quality:\footnote{
If a reference translation $r$ is also included in the input, then the quality score is called a reference-based metric.
}
\begin{align}
\widehat{\mathrm{y}} = f(s, t)
\end{align}
Quality estimation models are usually trained with the $\mathcal{L}_2$ loss between predicted and human assessments of quality
\citep{kocmi-etal-2022-findings,kocmi-etal-2024-error,lommel2014multidimensional}.
\begin{align}
\mathcal{L}_2(y, f(s, t))
\end{align}

The COMET model \citep{rei-etal-2020-comet,rei-etal-2022-cometkiwi} is one such quality estimation function $f$.
The model is based on encoding the source, translation, and potentially also the reference with a multi-layer, multilingual, transformer-based encoder.
Once the embeddings are computed, they are joined in a regression head that produces the final score.

For epistemic\footnote{\citet{zerva-etal-2022-disentangling} distinguish between epistemic uncertainty, meaning lack of model knowledge, and aleatoric uncertainty, meaning noise in data.} uncertainty estimation, we wish to have a predictor corresponding to a particular quality estimation model $f$ that predicts the magnitude of the individual sample-level error:
\begin{align}
\hat{e} = |f(s, t) - y|
\end{align}
The negative error then corresponds to confidence or certainty.

\subsection{Models}

We first introduce our instant confidence model and then describe two prior approaches for uncertainty-aware quality estimation.

\begin{figure}[t]
    \centering
    \includegraphics[width=1\linewidth]{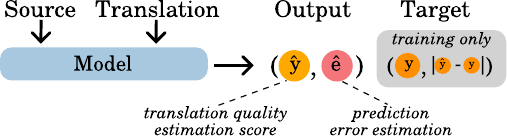}
    \caption{Architecture for uncertainty-aware quality estimation system based on COMET.}
    \label{fig:highlevel_lithium}
\end{figure}

\paragraph{Instant Confidence COMET.}
We propose modifying the quality estimation model to output both a quality estimate and an error estimate at the same time.
Thus, the model outputs $\langle\hat{y}, \hat{e}\rangle$, and during training we sum two MSE losses with importance of the second determined by hyperparameter $\beta$:
\begin{align}
\mathcal{L}_2 (y, \hat{y})+ \beta\cdot
\mathcal{L}_2 (|y-\hat{y}|, \hat{e})
\end{align}
Notably, we do not backpropagate loss through $\hat{y}$ in the second term, only through $\hat{e}$.
The architecture is illustrated in \Cref{fig:highlevel_lithium}.
During inference, the model has almost the same computational cost as an unmodified COMET model.

\paragraph{MC Dropout.}
\citet{glushkova-etal-2021-uncertainty-aware} elicit 
confidence scores
using Monte Carlo (MC) dropout.
This involves running the quality estimation model multiple times whilst introducing some randomness, in this case random dropout, and measuring the variance across runs in the output.
The underlying hypothesis is that the model outputs, even with dropout, will be the same for high-confidence samples but different for low-confidence samples.
While producing strong results, these methods are not practical for real applications because they can require up to $100$ model runs.

\paragraph{DUP.}
\citet{zerva-etal-2022-disentangling} train a separate secondary model that estimates the error of the original quality estimate. They call this Direct Uncertainty Prediction (DUP).
We consider two variants.
In the first, the uncertainty predictor does not know the original model prediction and in the second, as described in the previous work, the prediction is passed to the model.
Both variants come with a factor two computational overhead.
In addition, only the first variant is parallelizable because it does not depend on the original model's prediction.

\subsection{Results}
\label{sec:goal_instant_confidence_results}

We now evaluate the proposed uncertainty-aware quality estimation model and compare it to previous work.
We reproduce the previous works under the same conditions to have a level playing field for fair comparisons.

\paragraph{Setup.}
In line with previous works, we use the human-annotated segment-level data from WMT 2019 to 2022 for training \citep{barrault-etal-2019-findings,barrault-etal-2020-findings,akhbardeh-etal-2021-findings-FIXED,kocmi-etal-2022-findings} and reserve WMT 2023 \citep{kocmi-etal-2023-findings} for test set (116k segments across 8 language pairs: De$\leftrightarrow$En, Ja$\leftrightarrow$En, En$\leftrightarrow$Zh, Cs$\rightarrow$Uk, En$\rightarrow$Cs).\footnote{We do not use WMT 2024 \citep{kocmi-etal-2024-findings} due to the new annotation protocol which leads to lower correlations for all metrics.
}
We describe the general model technical details in Appendix \Cref{tab:model_details}.

\begin{table}[t]
\small
\centering
\begin{tabular}{l@{\hspace{1mm}}r@{\hspace{1mm}}c@{\hspace{2mm}}c@{}l@{}}
\toprule
& \bf Cost$\bm{\downarrow}$ & \bf Human$\bm{\uparrow}$ & \bf Error$\bm{\uparrow}$  \\
\midrule
Instant Confidence $\beta{=}0.25$\hspace{-2mm} & 1$\times$ & 0.316 & 0.222 \\
Instant Confidence $\beta{=}0.5$\hspace{-2mm} & 1$\times$ & 0.309 & 0.224 \\
Instant Confidence $\beta{=}0.75$\hspace{-2mm} & 1$\times$ & 0.326 & 0.228 & $\bigstar$ \\
Instant Confidence $\beta{=}1.0$\hspace{-2mm} & 1$\times$ & 0.330 & 0.207 \\
Instant Confidence $\beta{=}1.5$\hspace{-2mm} & 1$\times$ & 0.325 & 0.200 \\[0.5em]
No Confidence & 1$\times$ & 0.327 & - \\[0.5em]

MC Dropout (2) & 2$\times$ & 0.210 & 0.201 \\
MC Dropout (5) & 5$\times$ & 0.247 & 0.267 \\
MC Dropout (10) & 10$\times$ & 0.262 & 0.301 \\
MC Dropout (50) & 50$\times$ & 0.279 & 0.328 \\
MC Dropout (100) & 101$\times$ & 0.281 & 0.333 \\[0.5em]
MC Dropout$^\dagger$ (2) & 3$\times$ & 0.327 & 0.061 \\
MC Dropout$^\dagger$ (5) & 6$\times$ & 0.327 & 0.092 \\
MC Dropout$^\dagger$ (10) & 11$\times$ & 0.327 & 0.115 \\
MC Dropout$^\dagger$ (50) & 51$\times$ & 0.327 & 0.131 \\
MC Dropout$^\dagger$ (100) & 101$\times$ & 0.327 & 0.134 \\[0.5em]
DUP (parallelizable) & 2$\times$ & 0.327 & 0.135 \\  
DUP (sequential) & 2$\times$ & 0.327 & 0.216 \\
\bottomrule

\end{tabular}
\caption{Correlation (Pearson) of model scores with human scores (Human) and correlation of model error predictions (negative confidence) with true error (Error). Higher is better $\uparrow$ and lower is better $\downarrow$. MC Dropout$^\dagger$ uses dropout only for predicting the error and calculates the quality score without dropout.
}
\label{tab:mcdropout}
\end{table}

\paragraph{Pointwise confidence is cheap.}

We compare our approach with previous work in \Cref{tab:mcdropout}.
The $\beta$ hyperparameter allows for trade-off between correlation with human score and confidence correlation with true error.
For Monte Carlo dropout, as the number of runs is increased, the correlation between the model score variance (model confidence) and the true error increases.
However, this comes at the very steep cost of having to run the same model with dropout multiple times, which is not feasible in many quality estimation applications.
Moreover, using dropout during inference hurts the quality estimation performance, because part of the information flow in the network is obscured.
Increasing the number of runs only partially compensates for this drop; indeed, even with 100 runs, the score correlation remains lower than that obtained with a single run with no dropout (0.281 vs. 0.327).

Upon first glance, the solution appears to be to run Monte Carlo dropout 100 times to obtain a good uncertainty estimate and once without the dropout to also obtain a good quality estimation.
We show this approach as MC Dropout$^{\dagger}$ in \Cref{tab:mcdropout}.
However, due to the discrepancy between this model score and the uncertainty measure, the error correlation turns out to be very low, revealing this not to be a viable uncertainty prediction setup.

Lastly, we compare with DUP, which has the original (best) human correlation but still twice the cost of our approach and lower error correlation.

Overall, our instant confidence approach is the best in terms of cost and competitive in terms of human and confidence correlations.
In Appendix \Cref{fig:15c-expected_calibration_error} we see that on average 
when the quality estimation model has high confidence (low $\hat{e}$) then it is likely to have an accurate prediction (low $|\hat{y}-y|$).
Notably, the lowest confidence bin corresponds to samples where the quality estimation model is very incorrect ($|\hat{y}-y|>20$).

\section{Early-Exit COMET}
\label{sec:goal_earlyexit1}

COMET is a computationally expensive evaluation method. The $24$ transformer layers of the model are costly to compute. The idea of Early-Exit COMET is to use the embeddings from earlier layers to predict the final COMET score in advance. In addition, each layer predicts the error between its score estimate and the full COMET score. In this way, if the model has a confident estimate of the full COMET score after a few layers, then the evaluation can \emph{early-exit} and save on compute costs.

\begin{figure}[t]
    \centering
    \includegraphics[width=1\linewidth]{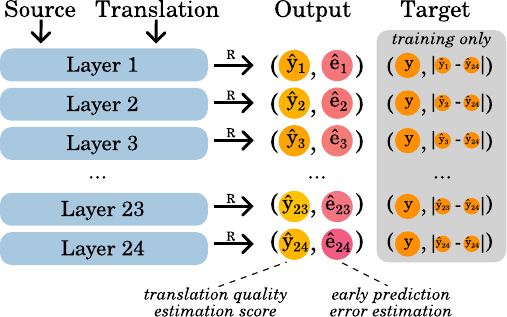}
    \caption{Architecture for confidence-aware (with respect to last layer) early-exit quality estimation system based on COMET.}
    \label{fig:highlevel_earlyexit}
\end{figure}

\subsection{Early-Exit COMET with Self-Confidence}
\label{sec:earlyexit_model}

We make two architectural changes to COMET to enable confidence-aware early-exit:
(1) predictions at each layer instead of just after the final layer, and
(2) self-confidence predictions that predict the error with respect to the final layer COMET score.\footnote{This is distinct from Instant Confidence COMET, where the predicted error was with respect to human scores.}
Importantly, we do not attempt to predict the error with respect to human scores because we wish to stop if we are confident that the final layer would also not produce a very different output.

Let $\mathrm{L}_i$ be the embedding after layer $i$, and let $\mathrm{R}$ be the regressor head on top of the final layer.
The final model prediction is then:
\begin{align}
\hat{y} = \mathrm{R}(\mathrm{L}_{|\mathrm{L}|}).
\end{align}
We wish to predict approximate scores $\hat{y}_i$ after each layer $i$.
Therefore, instead of training the evaluation model with the standard loss function,
\begin{gather}
\mathcal{L}_2 \left(y, \mathrm{R}(\mathrm{L_\mathrm{|L|})} \right),
\end{gather}
we apply the same regressor head to each layer and calculate the cumulative loss:
\begin{align}
\sum_{i=1}^{|\mathrm{L}|} \mathcal{L}_2 \left(y, \mathrm{R}(\mathrm{L_i}) \right).
\end{align}

To incorporate confidence predictions, we also increase the output dimensionality of the regressor head to two ($\mathrm{R}_{\mathrm{y}}$ and $\mathrm{R}_{\mathrm{e}}$).
The second output is used to estimate how far the (early) prediction is from the final prediction.
We refer to this as \textit{self-confidence}.
This gives an additional loss term for each layer:
\begin{align}
\mathcal{L}_2 \left(|\mathrm{R}_{y}(\mathrm{L}_i)-\mathrm{R}_{y}(\mathrm{L_\mathrm{|L|}})|\, , \,\mathrm{R}_{e}(\mathrm{L_i})\right).
\end{align}
The architecture is illustrated in \Cref{fig:highlevel_earlyexit}.

\begin{table}[t]
\newcolumntype{V}{>{\raggedleft\arraybackslash}p{6.8mm}}
\newcolumntype{Z}{>{\raggedleft\arraybackslash}p{8.3mm}}
\small
\setlength{\tabcolsep}{3pt}
\centering

\resizebox{\linewidth}{!}{
\begin{tabular}{llVVVVVVVZ}
\parbox[t]{2mm}{\multirow{11}{*}{\rotatebox[origin=c]{90}{\bf Layer}}}
& \multicolumn{7}{c}{\bf \color{green!40!black} Layers} 
& \multicolumn{2}{r}{\bf \color{purple!80!black} Human} \\
 &  & \tiny 01\,\,\,\, & \tiny 05\,\,\,\, & \tiny 09\,\,\,\, & \tiny 13\,\,\,\, & \tiny 17\,\,\,\, & \tiny 21\,\,\,\, & \tiny 24\,\,\,\,\\
& \tiny 01
& \cellcolor{green!40} 1.00 & \cellcolor{green!17} 0.30 & \cellcolor{green!14} 0.23 & \cellcolor{green!12} 0.17 & \cellcolor{green!12} 0.17 & \cellcolor{green!12} 0.15 & \cellcolor{green!12} 0.15 & \cellcolor{purple!4} 0.034
\\
& \tiny 05
& \cellcolor{green!17} 0.30 & \cellcolor{green!40} 1.00 & \cellcolor{green!38} 0.93 & \cellcolor{green!31} 0.72 & \cellcolor{green!30} 0.70 & \cellcolor{green!29} 0.66 & \cellcolor{green!28} 0.65 & \cellcolor{purple!19} 0.207
\\
& \tiny 09
& \cellcolor{green!14} 0.23 & \cellcolor{green!38} 0.93 & \cellcolor{green!40} 1.00 & \cellcolor{green!33} 0.78 & \cellcolor{green!32} 0.75 & \cellcolor{green!30} 0.70 & \cellcolor{green!30} 0.69 & \cellcolor{purple!20} 0.221
\\
& \tiny 13
& \cellcolor{green!12} 0.17 & \cellcolor{green!31} 0.72 & \cellcolor{green!33} 0.78 & \cellcolor{green!40} 1.00 & \cellcolor{green!39} 0.97 & \cellcolor{green!36} 0.87 & \cellcolor{green!35} 0.85 & \cellcolor{purple!25} 0.278
\\
& \tiny 17
& \cellcolor{green!12} 0.17 & \cellcolor{green!30} 0.70 & \cellcolor{green!32} 0.75 & \cellcolor{green!39} 0.97 & \cellcolor{green!40} 1.00 & \cellcolor{green!37} 0.91 & \cellcolor{green!36} 0.89 & \cellcolor{purple!26} 0.281
\\
& \tiny 21
& \cellcolor{green!12} 0.15 & \cellcolor{green!29} 0.66 & \cellcolor{green!30} 0.70 & \cellcolor{green!36} 0.87 & \cellcolor{green!37} 0.91 & \cellcolor{green!40} 1.00 & \cellcolor{green!40} 0.99 & \cellcolor{purple!28} 0.312
\\
& \tiny 24
& \cellcolor{green!12} 0.15 & \cellcolor{green!28} 0.65 & \cellcolor{green!30} 0.69 & \cellcolor{green!35} 0.85 & \cellcolor{green!36} 0.89 & \cellcolor{green!40} 0.99 & \cellcolor{green!40} 1.00 & \cellcolor{purple!28} 0.309
\\

\end{tabular}
}

\vspace{-2mm}
Early-Exit COMET

\vspace{-1mm}

\resizebox{\linewidth}{!}{
\begin{tabular}{llVVVVVVVZ}
\parbox[t]{2mm}{\multirow{11}{*}{\rotatebox[origin=c]{90}{\bf Layer}}}\\
 &  & \tiny 01\,\,\,\, & \tiny 05\,\,\,\, & \tiny 09\,\,\,\, & \tiny 13\,\,\,\, & \tiny 17\,\,\,\, & \tiny 21\,\,\,\, & \tiny 24\,\,\,\,\\
& \tiny 01
& \cellcolor{green!40} 1.00 & \cellcolor{green!14} 0.23 & \cellcolor{green!12} 0.17 & \cellcolor{green!10} 0.10 & \cellcolor{green!6} -0.01 & \cellcolor{green!4} -0.07 & \cellcolor{green!5} -0.06 & \cellcolor{purple!0} -0.033
\\
& \tiny 05
& \cellcolor{green!14} 0.23 & \cellcolor{green!40} 1.00 & \cellcolor{green!35} 0.86 & \cellcolor{green!29} 0.68 & \cellcolor{green!23} 0.48 & \cellcolor{green!19} 0.38 & \cellcolor{green!16} 0.28 & \cellcolor{purple!7} 0.064
\\
& \tiny 09
& \cellcolor{green!12} 0.17 & \cellcolor{green!35} 0.86 & \cellcolor{green!40} 1.00 & \cellcolor{green!37} 0.90 & \cellcolor{green!29} 0.68 & \cellcolor{green!25} 0.56 & \cellcolor{green!20} 0.41 & \cellcolor{purple!11} 0.116
\\
& \tiny 13
& \cellcolor{green!10} 0.10 & \cellcolor{green!29} 0.68 & \cellcolor{green!37} 0.90 & \cellcolor{green!40} 1.00 & \cellcolor{green!35} 0.86 & \cellcolor{green!31} 0.73 & \cellcolor{green!24} 0.53 & \cellcolor{purple!16} 0.176
\\
& \tiny 17
& \cellcolor{green!6} -0.01 & \cellcolor{green!23} 0.48 & \cellcolor{green!29} 0.68 & \cellcolor{green!35} 0.86 & \cellcolor{green!40} 1.00 & \cellcolor{green!39} 0.96 & \cellcolor{green!31} 0.73 & \cellcolor{purple!24} 0.264
\\
& \tiny 21
& \cellcolor{green!4} -0.07 & \cellcolor{green!19} 0.38 & \cellcolor{green!25} 0.56 & \cellcolor{green!31} 0.73 & \cellcolor{green!39} 0.96 & \cellcolor{green!40} 1.00 & \cellcolor{green!33} 0.80 & \cellcolor{purple!26} 0.283
\\
& \tiny 24
& \cellcolor{green!5} -0.06 & \cellcolor{green!16} 0.28 & \cellcolor{green!20} 0.41 & \cellcolor{green!24} 0.53 & \cellcolor{green!31} 0.73 & \cellcolor{green!33} 0.80 & \cellcolor{green!40} 1.00 & \cellcolor{purple!29} 0.327
\\

\end{tabular}
}

\vspace{-2mm}
Baseline COMET

\caption{Pearson correlations between intermediate layer outputs (green) and between intermediate layer outputs and humans (pink) for supervised Early-Exit as described in \Cref{sec:earlyexit_model} (left) and unsupervised Early-Exit based on standard COMET (right). See Appendix \Cref{10-eval_all_big} for detailed version.}
\label{10-eval_oxygen_hydrogen}

\vspace{-1mm}
\end{table}

\subsection{Results}

We now discuss our findings with Early-Exit COMET.
We use the same experimental setup as in \Cref{sec:goal_instant_confidence}.

\paragraph{Early layer scores.}
To measure the quality of the early layer scores, we calculate correlations with the final layer scores, as well as with human evaluations.
For comparison, we include the baseline COMET model, applying the final layer regressor to the intermediate embeddings to get intermediate scores.
In \Cref{10-eval_oxygen_hydrogen} we show that earlier layer scores of baseline COMET do not correlate strongly with final layer scores or with human judgments. 
However, with direct supervision at each layer, we see much better results for Early-Exit COMET.
At layer 5 we already see a correlation score of $0.65$ with the final layer and $0.207$ with human scores. By layer $13$, the correlation with human scores is $0.278$, comparable to the final layer.
We include a version of Early-Exit COMET with separate regression heads for each layer in Appendix \Cref{10-eval_all_big}, but we do not observe any improvements.

To measure the quality of the self-confidence error predictions, we plot the average predicted error versus the true error in Appendix \Cref{fig:13-plot_conf}. We also include correlation scores for selected layers showing the correlation between the predicted and true errors, e.g., $0.44$ for layer $9$ scores.
This enables early-exit decision making, which we introduce in the next section.

\subsection{Deciding When to Early-Exit}
\label{sec:faster_segment_qe}

In some cases, the Early-Exit COMET model is already close to the final assessment after a few layers (top row in \Cref{fig:14-plot_conf_individual}).
In these cases, we do not need to continue the computation if we are confident that the final outcome will be very close.
The per-layer confidences can inform this decision.

In \Cref{alg:threshold_algorithm}, we outline a simple heuristic that stops the Early-Exit COMET computation when the error prediction is low enough.
To evaluate this algorithm, we compare it with two baselines that do not use the confidence scores: (1) Constant-Exit (Appendix, \Cref{alg:constant_algorithm}) stops at a constant, pre-defined layer; and (2) Variance-Exit (Appendix, \Cref{alg:variance_algorithm}) stops when the variance of three consecutive predictions is under a chosen threshold.

\begin{figure}[t]
{
\small
\hrule
\vspace{1mm}
\textbf{Inputs}: Source $s$, translation $t$, threshold $\tau$ \\
\textbf{Output}: Quality estimate $\hat{y}$ 
\vspace{1mm}
\hrule
\vspace{1mm}
\begin{algorithmic}[1]

\State Compute $L_0(s, t)$
\For{$i \in 1\ldots |L|$}
    \State {Compute }$L_i(s,t)$ from $L_{i-1}(s,t)$ \Comment{next layer}
    \State $\hat{y}_i, \hat{e}_i \gets R(L_i)$ \Comment{apply regressor head}
    \State \textbf{if} \, $\hat{e}_i < \tau$ \, \textbf{then return} $\hat{y}_i$
    \Comment{early-exit}
\EndFor
\State \Return  $\hat{y}_{|L|}$ 
\end{algorithmic}
}
\vspace{1mm}
\hrule
\vspace{1mm}

\captionof{algorithm}{Confidence-Exit with Early-Exit COMET.}
\label{alg:threshold_algorithm}
\vspace{-2mm}
\end{figure}

\begin{figure}[t]
    \centering
    \includegraphics[width=\linewidth]{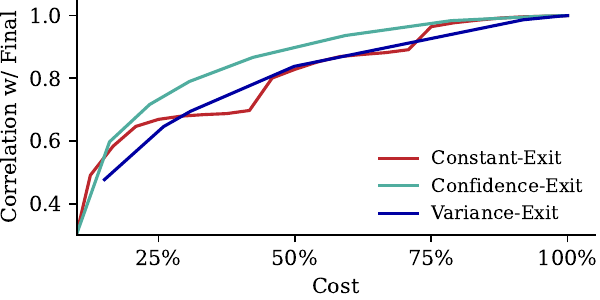}

    \vspace{-3.2mm}
    \includegraphics[width=\linewidth]{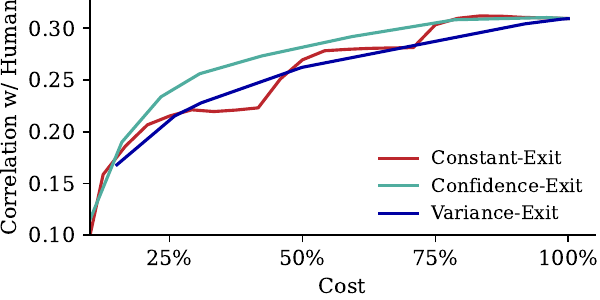}

    \vspace{-2mm}
    \caption{
    Quality estimation correlation with last layer (final prediction, top) and with human scores (bottom) for heuristic-based early-exit COMET.
    }
    \label{fig:20-early_exit_alg}
    \vspace{-1mm}
\end{figure}

\begin{figure*}[t]
{
\small
\hrule
\vspace{1mm}
\textbf{Inputs}: Early-Exit model with predictions $\hat{y}_l$ and confidences (error estimates) $e_l$ for layer $l \in [1, |L|]$, translation candidates $\mathcal{C}$, exploration-exploitation hyperparameter $\gamma$, total evaluation budget $B$ \\
\textbf{Output}: final translation candidate $c^* \in \mathcal{C}$  
\vspace{1mm}
\hrule
\vspace{1mm}
\begin{algorithmic}[1]

\State $\sigma \gets e \cdot \sqrt{\frac{\pi}{2}}$ \Comment{Rescale the absolute error estimates (MAE) to standard deviation estimates}
\State $S_{\mathcal{C}} \gets \{ (\hat{y}_1(c), \sigma_1(c)) | c \in \mathcal{C}\}$ \Comment{Calculate and cache first layer COMET for all candidates}
\State $\hat{B} \gets |\mathcal{C}|$ \Comment{Initialize running evaluation costs}
\State $\mathcal{C}' \gets \mathcal{C}; \quad l(c) \gets 1 \ \forall c \in \mathcal{C}'$ \Comment{Initialize remaining candidates and explored layers}
\While{$\hat{B}<B$ \textbf{and} $\mathcal{C}' \neq \emptyset$}
    \State $\mathrm{UCB}(c) \gets \hat{y}_{l(c)}(c) + \gamma \sigma_{l(c)}(c) \quad \forall c \in \mathcal{C}'$ \Comment{Calculate UCB for all (remaining) candidates}
    \State $c^* \gets \argmax_{c \in \mathcal{C}'} \mathrm{UCB}(c)$ \Comment{Choose candidate with highest UCB}
    \State $l(c^*) \gets l(c^*)+1; \quad \hat{B} \gets \hat{B}+1$
    \State $S_{\mathcal{C}} \gets S_{\mathcal{C}} \cup (\hat{y}_{l(c^*)}(c^*), \sigma_{l(c^*)}(c^*))$ \Comment{Calculate next layer COMET score for $c^*$}
    \If{$l(c^*) = |L|$}
    \State $\mathcal{C}' \gets \mathcal{C}' \setminus \{c^*\}$ \Comment{Remove $c^*$ if fully evaluated}
    \EndIf

\EndWhile
\Comment{Return best candidate based on the most advanced}
\State \Return  $c^* \gets \arg\max_{c \in \mathcal{C}}(\hat{y}_{l(c)}(c)) \quad$ 
\hfill (highest layer) predictions for each candidate.
\end{algorithmic}
}
\vspace{1mm}
\hrule
\vspace{1mm}

\captionof{algorithm}{Upper Confidence Bound (UCB) Bandit for selecting a high scoring translation candidate from a pool of candidates using Early-Exit COMET.}
\label{alg:ucb_algorithm}
\vspace{-5mm}
\end{figure*}

\paragraph{Results.}

\Cref{fig:20-early_exit_alg} shows the performance of \Cref{alg:threshold_algorithm} versus the baselines. We plot the correlation with final layer scores and human judgments for different budgets (total computation costs). The cost is relative to calculating the full COMET score for all inputs ($100\%$).
We vary the algorithm thresholds to explore different budgets.

We see that the confidence-based early-exit algorithm outperforms variance-exit and constant-exit.
This shows that the confidence outputs at early layers are crucial for enabling early-exit.
With only half of the compute, we see only a small performance drop from the full to half compute: ($0.309{\rightarrow}0.292$) for human scores.

\begin{figure*}[ht]
    \centering
    Beam-Search\\
    \includegraphics[width=0.49\linewidth]{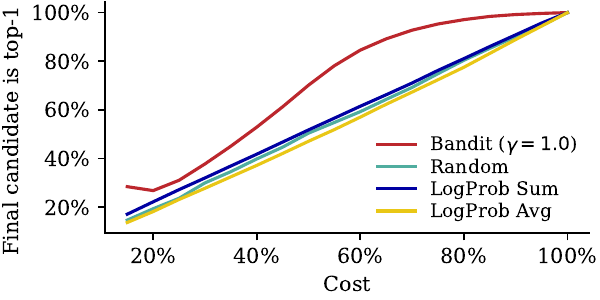}
    \hfill
    \includegraphics[width=0.49\linewidth]{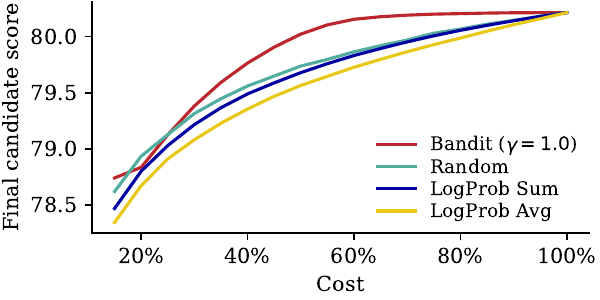}
    
    Sampling\\
    \includegraphics[width=0.49\linewidth]{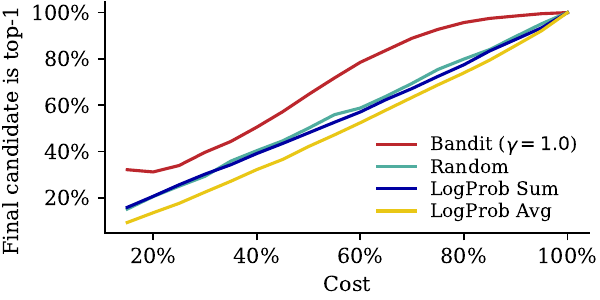}
    \hfill
    \includegraphics[width=0.49\linewidth]{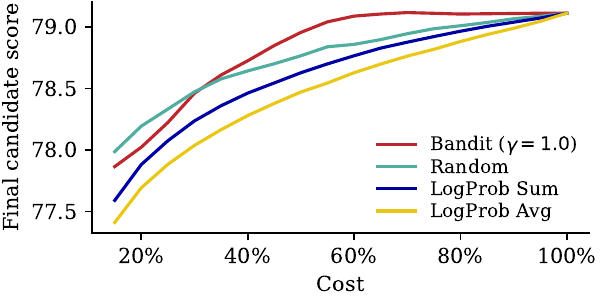}

    \vspace{-2mm}
    \caption{Quality of the candidates returned by the Upper Confidence Bound bandit. Quality is measured in terms of the average final candidate score and the proportion to top-1 candidates selected. We plot these measures for various evaluation budgets. Cost (or budget) is given relative to calculating the full COMET scores for all candidates ($100\%$).
    See results in tabular form in Appendix \Cref{tab:bandit}.
    }
    \label{fig:bandit}
\end{figure*}

\section{Early-Exit COMET for Reranking}
\label{sec:goal_earlyexit2}

In addition to enhancing the speed of quality estimation, the early-exit model can be applied to reranking. 
In this setup, a machine translation model generates a set of candidate translations, $\mathcal{C}$, for a source sentence, and
the objective of reranking is to identify the best candidate from $\mathcal{C}$.

Reranking with quality estimation has been shown to improve translation quality \citep{freitag-etal-2022-high} and, as one might expect, the larger the initial pool of candidates, the higher the final translation quality \citep{vernikos-popescu-belis-2024-dont}.
However, running a quality estimation model on a large number of candidates, e.g. $|\mathcal{C}| > 100$, can be prohibitively expensive.

To lower these compute costs, we turn to Early-Exit COMET and rely on its accurate early-layer predictions to make reranking more efficient.
The idea is to calculate more accurate (and more expensive) scores only for the most promising candidates based on the less expensive early-layer scores.

Each early-exit output, $\hat{y}_1, \hat{y}_2, \ldots, \hat{y}_{|L|}$, has an associated runtime cost: $\mathrm{cost}(\hat{y}_i)$. For Early-Exit COMET, we can take $\mathrm{cost}(\hat{y}_i) = i$ since each layer has the same computation cost.
Note that costs are not additive; for example, once the layer $3$ score is calculated, all later layers can use the layer $3$ embeddings. So for $i>3$, we will now only accumulate the additional cost of $\mathrm{cost}(\hat{y}_{i}) - \mathrm{cost}(\hat{y}_{3})$.

Ultimately, we must select a final candidate
and our goal is to strike a balance between the total computation cost and finding the candidate with the highest $\hat{y}_{|L|}$ score.

\subsection{Upper Confidence Bound Bandit}
In the multi-armed bandit problem, a decision-maker must repeatedly choose between several actions, ``arms'', with unknown reward distributions.
The goal is to maximize the cumulative reward earned over time by balancing exploration (trying out different arms to learn their rewards) and exploitation (choosing the arm believed to be the best), which can be done with the upper confidence bound bandit \citep{auer2002using}.
The algorithm computes an upper confidence bound for the estimated reward for each arm, selecting the arm with the highest bound in each round.
This approach encourages the algorithm to explore less-pulled arms (with larger error estimates) while exploiting those with higher estimated rewards.

In our context, the ``arms'' are the translation candidates $\mathcal{C}$, and pulling an arm corresponds to calculating an additional quality estimation model layer.
Since the predictions improve with more layers (\Cref{10-eval_oxygen_hydrogen}), each ``pull'' of the ``arm'' improves our estimate of the reward for the given candidate.
For each candidate, we always consider the reward of the highest layer explored thus far and its associated confidence score.
The computation budget determines the number of pulls available. When the budget runs out, we pick the candidate with the highest reward estimate, ignoring the associated confidence scores. 
A more precise formulation is provided in \Cref{alg:ucb_algorithm}.
The $\gamma$ controls the balance between exploration and exploitation. We use $\gamma=1$ for most experiments, but an ablation study can be found in \Cref{sec:bandit_ablations}.

\paragraph{Setup.}
To evaluate our upper confidence bound bandit with Early-Exit COMET, we again use the WMT 2023 test set \citep{kocmi-etal-2023-findings}, including 8 language pairs\footnote{Due to the large size, we use a random subset of the test data with $2000$ source examples for each language pair.}. We generate translation candidates using NLLB \citep{nllb2022}, with 200 candidates per segment. We generate candidates via multinomial sampling (across the whole vocabulary) and via beam search separately and report results for both.

\paragraph{Baselines.}
We compare against a \textit{Random} baseline, where we select a random subset of candidates, calculate the full COMET scores for this subset and select the candidate with the highest score. The size of the subset is proportional to the budget.
In addition to the random baseline, we also show results for two baselines based on the log probability scores (logprobs) of the translation model (NLLB).
In contrast to the random baseline, we select the subset based on the logprobs of the candidate tokens. We take the highest-scoring candidates in terms of the average logprob score (\textit{LogProb Avg}) or the sum of logprob scores (\textit{LogProb Sum}), and calculate the full COMET scores for this subset.

\subsection{Results}

\Cref{fig:bandit} shows the quality of the bandit output as we increase the evaluation budget.
We plot both the average score of the final candidates and the rate at which the top-1 (best full COMET score) candidate is selected. 

Surprisingly, the logprob-based baselines underperform the random baseline in almost all settings, in both metrics, and for all budgets.
We hypothesize that this is due to lower diversity in the selected candidates subset. 
In contrast, the bandit outperforms the random baseline in almost all scenarios and for all budgets.
In particular, with $60\%$ of the compute budget, there is almost no drop in the translation quality.
This indicates that (1) the Early-Exit COMET scores and error predictions are valuable and (2) the upper confidence bound bandit makes efficient use of these estimates.

For more detailed results on the upper confidence bound bandit, please see the ablation studies in \Cref{sec:bandit_ablations}.

\section{Deferring to Humans}
\label{sec:deferral}

In this section, we use the confidence scores from 
{Instant Confidence COMET}
to decide whether we can rely directly on the COMET score, or whether a translation should instead be scored by a human annotator.

\paragraph{Methods.}
As the simplest baseline, we consider randomly selecting examples for human annotation.
For a stronger baseline, as used in \citet{zouhar-etal-2025-ai}, we select the examples with the lowest scores for human annotation.
This can be done with existing metrics.
We compare this with our approach of human-annotating examples 
with the lowest confidence scores (highest predicted error).
Finally, for both the low-score and low-confidence approaches, we also consider oracle versions where we prioritize examples with low human scores and high absolute error respectively.

\paragraph{Setup.}
We analyze deferral rates ranging between 0\% and 100\%.
To measure the success of the combined human-metric annotation, we look at how the final ranking of the machine translation model changes degrades as we score more of the data with the automated metric.
We quantify this by the system-level Spearman correlation between ranking based on the combined annotation approach and 100\% human annotation.
We macro-average results across all considered languages in the dataset.

\begin{figure}[t]
    \centering
    \includegraphics[width=1\linewidth]{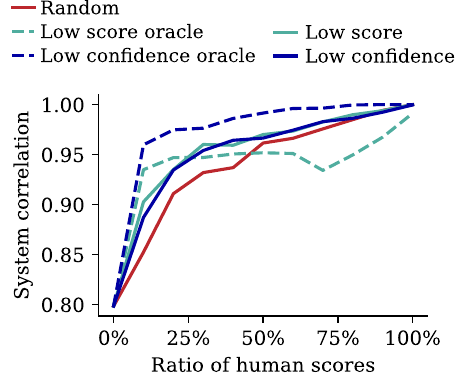}
    \caption{
    System-level correlation (macro Spearman) of a combined human-metric annotation with part of segments being annotated by humans. At 0\%, all segments are scored by the metric. At 100\%, all segments are scored by humans.}
    \label{fig:deferral_correlation}
\end{figure}

\paragraph{Results.}
The results are shown in \Cref{fig:deferral_correlation}.
Our method outperforms random selection, though is on par with
deferring low-scoring examples. 
However, lower scores correlate (0.243 macro Pearson) with the number of words in the translation.
The length of the translation affects the human annotation time \citep{zouhar-etal-2025-ai} and thus e.g. 20\% deferral with this approach might use more annotation budget than 20\% deferral with random selection.
On the other hand, low confidence does not correlate meaningfully (-0.087 macro Pearson) with the translation length, thus making it a safer option.

Finally, we note that there is a large gap between using the predicted confidence scores and the oracle confidence.
This means, that improvements in estimating the automated metric error can lead to further gains in deferral efficiency.

\section{Related Work}

We now position our contributions in relation to other works on uncertainty-aware quality estimation, faster inference, and reranking.

\paragraph{Model uncertainty.}
Quantifying uncertainty in learned automated machine translation metrics was first proposed by \citet{glushkova-etal-2021-uncertainty-aware}, who use the variance from ensembles or Monte Carlo dropout as a measure of uncertainty.
A different approach by \citet{zerva-etal-2022-disentangling} introduces a secondary confidence estimation model to complement the original quality estimation model.\footnote{
They also propose an approach whereby one model predicts both quantities, similar to our instant confidence.
See \Cref{sec:fail_hts} documenting our attempts at replicating this model.
}
Instead, we optimize a single model that produces two scores: a quality estimate and a measure of uncertainty.
We optimize this model jointly in each training step, optionally optimizing for predictions at each layer and optionally with respect to the last layer's prediction.
Finally, we note that the focus of \citet{zerva-etal-2022-disentangling} is on analyzing the source of uncertainty, while our focus is on using uncertainty to make quality estimation more reliable and efficient.
We direct the reader to the work of \citet{lahlou2023deupdirectepistemicuncertainty} for more theoretical yet comprehensive treatment of directly estimating model confidence.

\paragraph{Faster quality estimation.}
Multiple previous works investigated improving the efficiency of calculating trained metrics.
For large scales, \citet{rei-etal-2022-searching} use length-batching to speed up inference.
At the same time, they statically prune the quality estimation model, which is similar to our constant-exit approach.
\citet{cheng2024bayesianoptimizationapproachmachine} use a smaller baseline language model rather than the default XLM-Roberta for COMET.
\citet{zouhar-etal-2024-pitfalls} find that simply quantizing the model to half precision has almost no effect on the final quality estimation performance while halving the compute costs.
\citet{gowda-etal-2023-cometoid,gowda-etal-2024-pymarian} port COMET to a faster inference engine for massive speed gains.
\citet{larionov2024xcometlitebridginggapefficiency} explore pruning, distillation, and quantization for a very large quality estimation model, xCOMET \citep{guerreiro-etal-2024-xcomet}.
All of these approaches are orthogonal to our method
and could be used in combination.

\paragraph{Early-Exit.}
Many works have explored intermediate model predictions \citep[][inter alia]{liu-etal-2019-linguistic,belrose2023elicitinglatentpredictionstransformers}.
However, the key ingredient is to know when to stop the computation, such as at a particular Transformer block layer \citep{bertpatience}, but these methods are largely applicable to classification tasks.
\citet{xin-etal-2021-berxit} propose \textit{learning-to-exit}, which we loosely follow in our work.
However, instead of predicting a probability of success, we predict the absolute error of the model, which is directly interpretable.

\paragraph{Reranking.}
Reranking improves translation quality \citep{freitag-etal-2022-high}, but scoring large candidate sets is computationally expensive. One common approach is minimum Bayes risk (MBR) decoding \citep[MBR;][]{eikema-aziz-2020-map}, which selects the translation candidate with the lowest expected risk. Recent work has made MBR more efficient \citep{cheng-vlachos-2023-faster,deguchi-etal-2024-centroid,trabelsi2024efficientminimumbayesrisk,vamvas-sennrich-2024-linear}, including methods that pre-select candidates with cheaper, noisier scoring functions \citep{fernandes-etal-2022-quality, eikema-aziz-2022-sampling}.

Other approaches improve efficiency through token-level reranking \citep{singhal-etal-2023-eel} or by framing reranking as a Bayesian Optimization problem, where a cheaper scoring model assists in identifying high-quality candidates before applying the more expensive scoring model \citep{cheng2024bayesianoptimizationapproachmachine}.

\section{Conclusion}

We introduced three approaches to improve the efficiency of quality estimation.
Our instant confidence Early-Exit COMET achieves comparable performance to Monte Carlo dropout methods while drastically lowering computational overhead.
Combining our model with a simple early-exit strategy, we can compute comparable quality estimation scores without having to compute the full quality estimation model.
Combining Early-Exit COMET with an upper confidence bound bandit, we speed up candidate reranking for machine translation by a factor of almost $2$ with negligible impact on translation quality.
Finally, the confidence scores can also inform which translations to human-annotate.

\paragraph{Recommendations.}
Based on our findings, we offer the following practical advice:
\begin{itemize}[topsep=0mm]
\item When quality estimation is part of a more complex decision process, we recommend using instant confidence-aware COMET to provide additional information on the credibility of decisions.
\item For very large-scale quality estimation use cases with limited compute budget, we recommend Early-Exit COMET with Confidence-Exit.
\item For reranking with very large candidate pools, we recommend the upper confidence bound bandit to reduce the number of scored candidates.
\item When human-annotating only a portion of translations, prioritize those with low Instant Confidence COMET confidence.
\end{itemize}

\paragraph{Future work.}
Reranking can also be combined with beam search to improve quality at generation time. Future work could apply our ideas to improve reranking efficiency for model generation. 
Moreover, in \Cref{sec:goal_partial} we describe a prototype of Partial COMET, a quality estimation model that is able to robustly evaluate incomplete generations.
This can be used to prune incomplete candidates, thus saving unnecessary computation of the very expensive generative model.

\section*{Limitations}

Regarding the results in \Cref{tab:mcdropout}, it is possible that there is a causal trade-off between the two correlation scores, the correlation with human scores and the correlation with the true error.
For example, it could be easier to predict the confidence of a model that performs worse.
However, making stronger claims would require a more thorough mathematical treatment, which is outside of the focus of this work.

\section*{Ethics Statement}
Data used in this paper was collected by previous works.
The authors foresee no ethical problems.

\section*{Acknowledgments}

We thank the EAMT committee for sponsoring this research. 
We thank the Vector Stiftung for supporting Béni Egressy's work.
Part of this work received support from the European Union’s Horizon research and
innovation programme under grant agreement No 101135798, project Meetween (My Personal AI Mediator for Virtual MEETtings BetWEEN People).
This research has been funded in part by a Swiss National Science Foundation award (project 201009) and a Responsible AI grant by the Haslerstiftung.

\bibliography{misc/anthology.min.bib,misc/bibliography.bib}
\bibliographystyle{misc/acl_natbib}

\clearpage

\appendix

\section{Model Architecture and Training Details}
\label{app:model_details}

\Cref{tab:model_details} provides details of the model architecture and training hyperparameters that were used to train the COMET models in this paper.

Unless otherwise specified, the configuration of \href{https://github.com/Unbabel/COMET/blob/master/configs/models/referenceless_model.yaml}{referenceless COMET} is used.
Our models are largely compatible with the upstream COMET repository and can be reproduced based on our code.
We run all our experiments on Nvidia A100 (40GB) GPUs, taking about 8 hours to train a single model (2.2GB) for 5 epochs.
For each setting, we train a single model and report its performance.
In total, all experiments, including preliminary ones, amounted to approximately 20$\times$8 hours = 160 hours of compute on the aforementioned GPU.
We base our experiments on a modified COMET v2.2.4 codebase with other package versions listed as dependencies in this version.

\begin{table}[htbp]
\small
\centering
\begin{tabular}{ll}
\toprule
Encoder & xlm-roberta-large (24 layers) \\
Embeddings & Layerwise attention \& CLS \\
Encoder frozen & 30\% of first epoch \\
Regression head & $(4{\times}768)\times 2048{\times}1024 \times (1$ or $2)$\\
Optimizer & AdamW \\
Learning rate & $1.5\times10^{-5}$, encoder $10^{-6}$ \\
Batch size &  256 (simulated) \\
Loss & MSE for both targets \\
Training epochs & 5 \\
\bottomrule
\end{tabular}
\caption{Model architecture and training details. }
\label{tab:model_details}
\end{table}

\section{Replicating HTS model}
\label{sec:fail_hts}
\citet{zerva-etal-2022-disentangling} propose a quality estimation model that outputs a distribution by predicting its mean and variance.
Loosely, this can be interpreted as the score and confidence prediction.
However, no public model is available and we have been unable to reproduce the model based on the \href{https://github.com/deep-spin/uncertainties_MT_eval}{publicly available code}.
Upon making the changes in the code that make the codebase compatible with up-to-date packages we did train a model with the \texttt{hts} loss, though the resulting human and error correlations were only 0.247 and 0.206, respectively.
Because this strays far from the original reported results, we attribute this to reproducibility failure as opposed to failure of the method.
\citet{zouhar-etal-2024-pitfalls} already show that the differences in COMET codebase versions can cause large discrepancies in COMET model behavior.
While our work uses the latest COMET v2.2.4, the work of \citet{zerva-etal-2022-disentangling} used COMET v1.0.0rc4.

\section{Additional Plots for Instant Confidence COMET}
\label{app:confidence_add_plots}

\begin{figure}[h]
    \centering
    \includegraphics[width=1\linewidth]{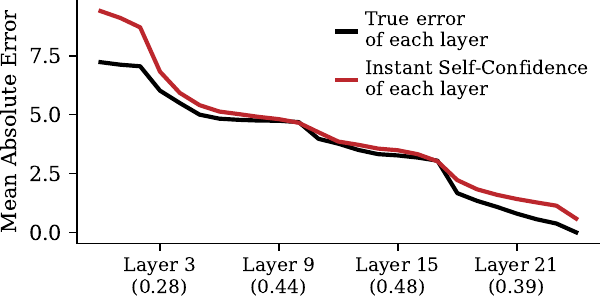}
    \caption{Correspondence of true and instant self-confidence. Correlations in brackets are Pearson correlation for each layer.}
    \label{fig:13-plot_conf}
\end{figure}

\begin{figure}[h]
    \centering
    \includegraphics[width=\linewidth]{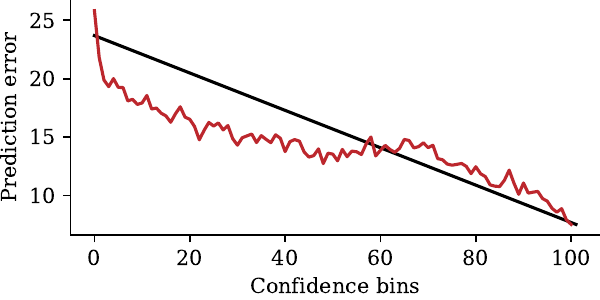}
    \caption{Calibration of predicted quality estimation model confidence based on 100 confidence bins (x-axis) and mean true absolute error of the prediction in each bin (y-axis).
    }
    \label{fig:15c-expected_calibration_error}
\end{figure}

We provide additional plots showing the quality of the Instant Confidence COMET error predictions. In \Cref{fig:13-plot_conf} we plot the average predicted error versus the true error to indicate average alignment. We also include correlation scores for selected layers showing the correlation between the predicted and true errors, e.g., $0.44$ for layer $9$ scores.
In \Cref{fig:15c-expected_calibration_error} we plot the average error for different confidence bins based on Instant Confidence. We see that the true error decreases as the predicted confidence increases. The plot also indicates that the score predictions with the highest and lowest true errors are reliably identified by the predicted instant confidence values.

\section{Baseline Early Exit Algorithms}
\label{app:early_exit_algorithms}

Below we provide pseudocode for the early-exit algorithms that we use as baselines to compare with Confidence-Exit. These baselines do not use the confidence scores of Early-Exit COMET.

\Cref{alg:constant_algorithm} exits the COMET evaluation after a fixed, predefined layer, exit layer $k$. 

\Cref{alg:variance_algorithm} exits the COMET evaluation when Early-Exit COMET scores from three consecutive layers are close to each other, or more precisely, have variance below a chosen threshold $\tau$.

\begin{figure}[ht!]
{
\small
\hrule
\vspace{1mm}
\textbf{Inputs}: Source $s$, translation $t$, exit layer $k$ \\
\textbf{Output}: Quality estimation $\hat{y}$ 
\vspace{1mm}
\hrule
\vspace{1mm}
\begin{algorithmic}[1]

\State Compute $L_0(s, t)$
\For{$i \in 1\ldots k$}
    \State {Compute }$L_i(s,t)$ from $L_{i-1}(s,t)$ \Comment{next layer}
\EndFor    
\State $\hat{y}_k, \hat{e}_k \gets R(L_k)$ \Comment{apply regressor head}
\State \Return  $\hat{y}_{k}$ 
\end{algorithmic}
}
\vspace{1mm}
\hrule
\vspace{1mm}

\captionof{algorithm}{Constant-Exit with Early-Exit COMET.}
\label{alg:constant_algorithm}

\end{figure}

\begin{figure}[ht!]
{
\small
\hrule
\vspace{1mm}
\textbf{Inputs}: Source $s$, translation $t$, threshold $\tau$ \\
\textbf{Output}: Quality estimation $\hat{y}$ 
\vspace{1mm}
\hrule
\vspace{1mm}
\begin{algorithmic}[1]

\State Compute $L_0(s, t)$
\For{$i \in 1\ldots |L|$}
    \State {Compute }$L_i(s,t)$ from $L_{i-1}(s,t)$ \Comment{next layer}
    \State $\hat{y}_i, \hat{e}_i \gets R(L_i)$ \Comment{apply regressor head}
    \State \textbf{if} $\mathrm{Var}[\hat{y}_{i-2:i}]{<} \tau$ \textbf{then return} $\hat{y}_i$
    \Comment{early-exit}
\EndFor
\State \Return  $\hat{y}_{|L|}$ 
\end{algorithmic}
}
\vspace{1mm}
\hrule
\vspace{1mm}

\captionof{algorithm}{Variance-Exit with Early-Exit COMET.}
\label{alg:variance_algorithm}

\end{figure}

\section{Upper Confidence Bound Bandit Ablations}
\label{sec:bandit_ablations}

We now describe two variations to the Upper Confidence Bound Bandit algorithm described in \Cref{sec:goal_earlyexit2}: (1) starting at later layers, and (2) balancing exploration and exploitation.

\begin{figure*}[t]
    \centering
    Beam-Search\\
    \includegraphics[width=0.49\linewidth]{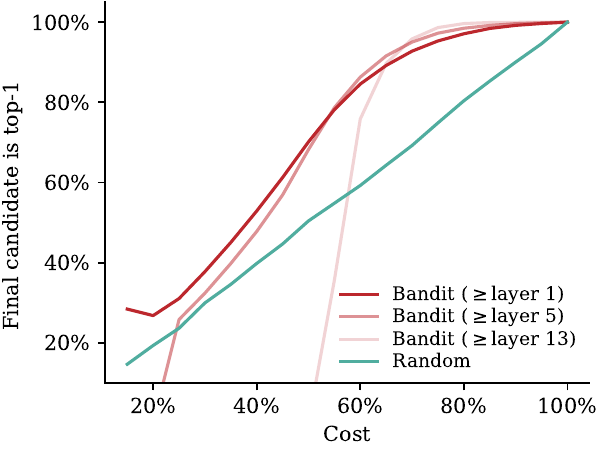}
    \hfill
    \includegraphics[width=0.49\linewidth]{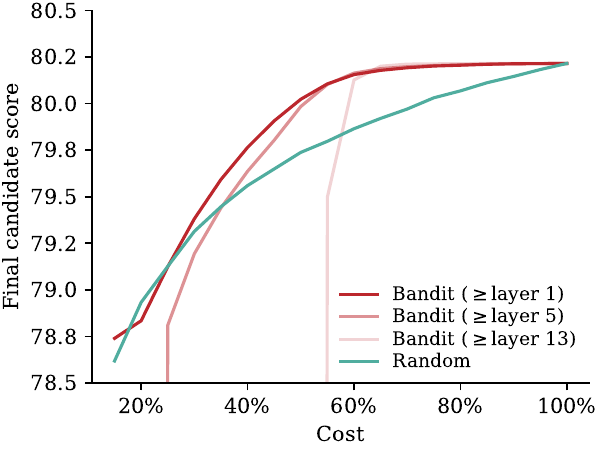}
    
    Sampling\\
    \includegraphics[width=0.49\linewidth]{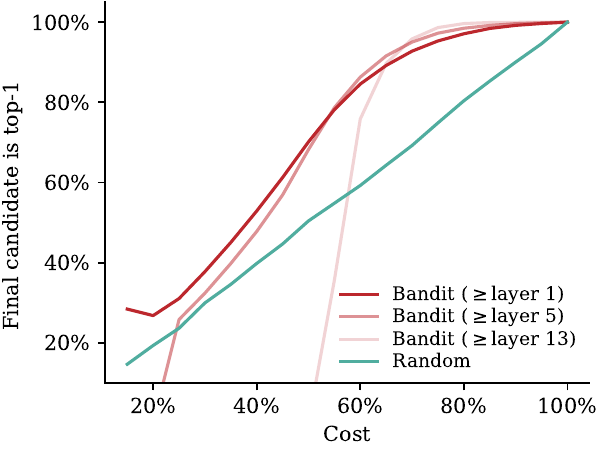}
    \hfill
    \includegraphics[width=0.49\linewidth]{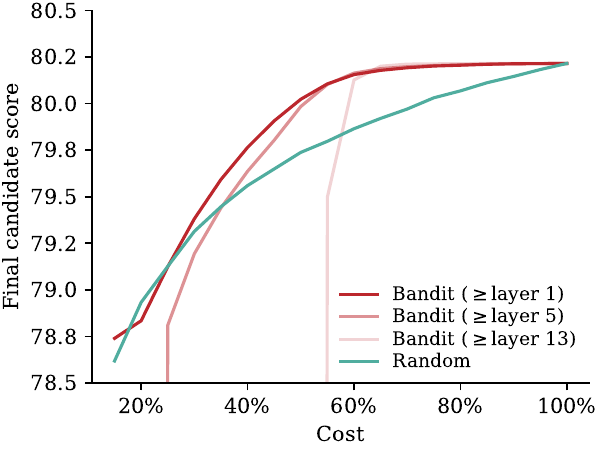}
    
    \caption{
    Ablation for the Upper Confidence Bound bandit by forcefully computing the first few layers. Quality is measured in terms of the average final candidate score and the proportion to top-1 candidates selected. We plot these measures for various evaluation budgets. Cost (or budget) is given relative to calculating the full COMET scores for all candidates ($100\%$).
    }
    \label{fig:bandit_ablation_start}
\end{figure*}

\begin{figure*}[t]
    \centering
    Beam-Search\\
    \includegraphics[width=0.49\linewidth]{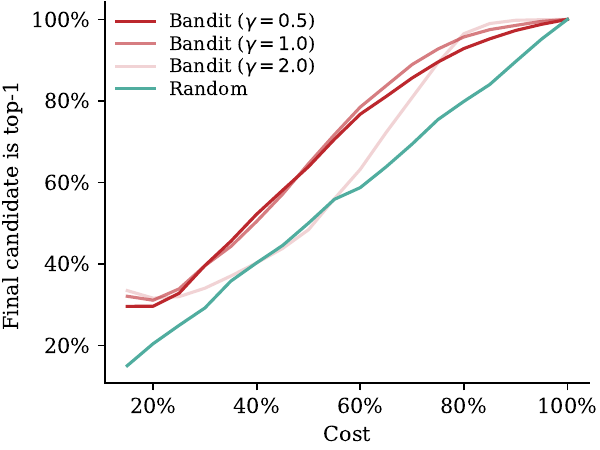}
    \hfill
    \includegraphics[width=0.49\linewidth]{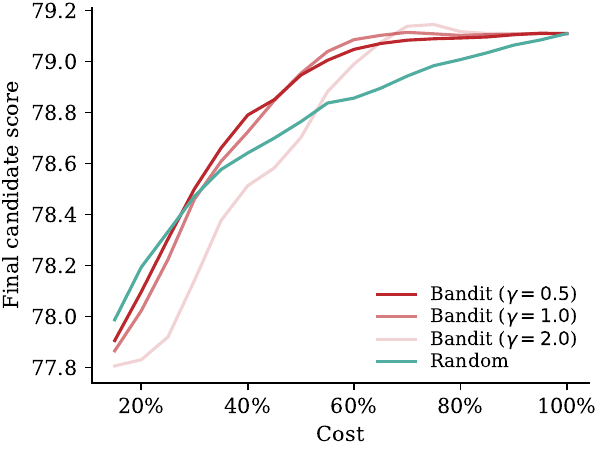}
    
    Sampling\\
    \includegraphics[width=0.49\linewidth]{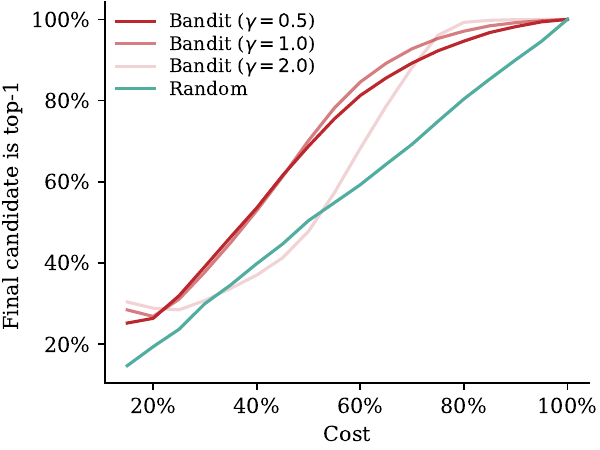}
    \hfill
    \includegraphics[width=0.49\linewidth]{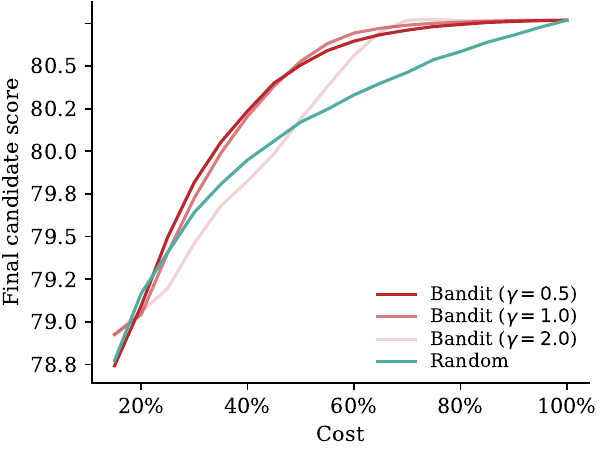}
    
    \caption{
    Ablation for the Upper Confidence Bound bandit with changing $\gamma$ (exploitation-exploration trade-off). With higher $\gamma$, the algorithm explores even otherwise low-scoring candidates. Quality is measured in terms of the average final candidate score and the proportion to top-1 candidates selected. We plot these measures for various evaluation budgets. Cost (or budget) is given relative to calculating the full COMET scores for all candidates ($100\%$).
    }
    \label{fig:bandit_ablation_gamma}
\end{figure*}

\paragraph{Starting at Different Layers.}
Given the significant jump in COMET score accuracy within the first few layers of the COMET model (\Cref{10-eval_oxygen_hydrogen}), we decided to explore initializing the algorithm with different starting layers. This carries a higher initial cost as we have to run the first few layers of the COMET model for all candidates but could lead to better-informed exploration with the remaining budget.
The results in \Cref{fig:bandit_ablation_start} show that this leads to only marginal improvements.
\paragraph{Exploration-Exploitation Tradeoff.}
The heart of the multi-armed bandit problem is the exploration-exploitation trade-off.
In our algorithm this trade-off is controlled by the hyperparameter $\gamma$.
The higher one chooses $\gamma$, the higher the Upper Confidence Bound scores for uncertain candidates will be, and therefore the more likely the algorithm will be to explore many candidates.
On the other hand, a low $\gamma$ will lead the algorithm to go deep with the most promising candidates, i.e., those with the highest estimated scores. 
We provide results for two different values for $\gamma$ in \Cref{fig:bandit_ablation_gamma}, which shows that the default choice is likely the most apt.

\paragraph{Distribution of Max. Layers Calculated.}
We analyze the order in which the layers were calculated by the UCB bandit algorithm by taking snapshots of the max. layer calculated across all candidates at $5\%$ budget increments, from $5\%$ to $100\%$. We provide plots for the default UCB bandit hyperparameter values (start layer = 1, $\gamma = 1.0$) on the test set with the candidates generated via sampling. 

At the start (budget, $B=5\%$), only the first layer COMET scores are explored for (almost) all candidates, and by the end (budget, $B=100\%$) the full COMET scores are calculated for all candidates.
However, the distributions in between reveal that the bandit often explores the candidates up until certain modal layers. These can be clearly seen (e.g. for budget $B=50\%$) to be layers $3$, $4$, $11$, $18$, and $24$. Looking at \Cref{fig:13-plot_conf}, it can be seen that these values correspond to layers where the predicted error rates (uncertainties) drop significantly, meaning that the evaluation model is now more certain about these scores relative to scores for other candidates. This explains why the UCB bandit algorithm would then prefer to explore less certain candidates with higher UCB scores.

\begin{figure*}[t]
    \centering
    \includegraphics[width=\linewidth]{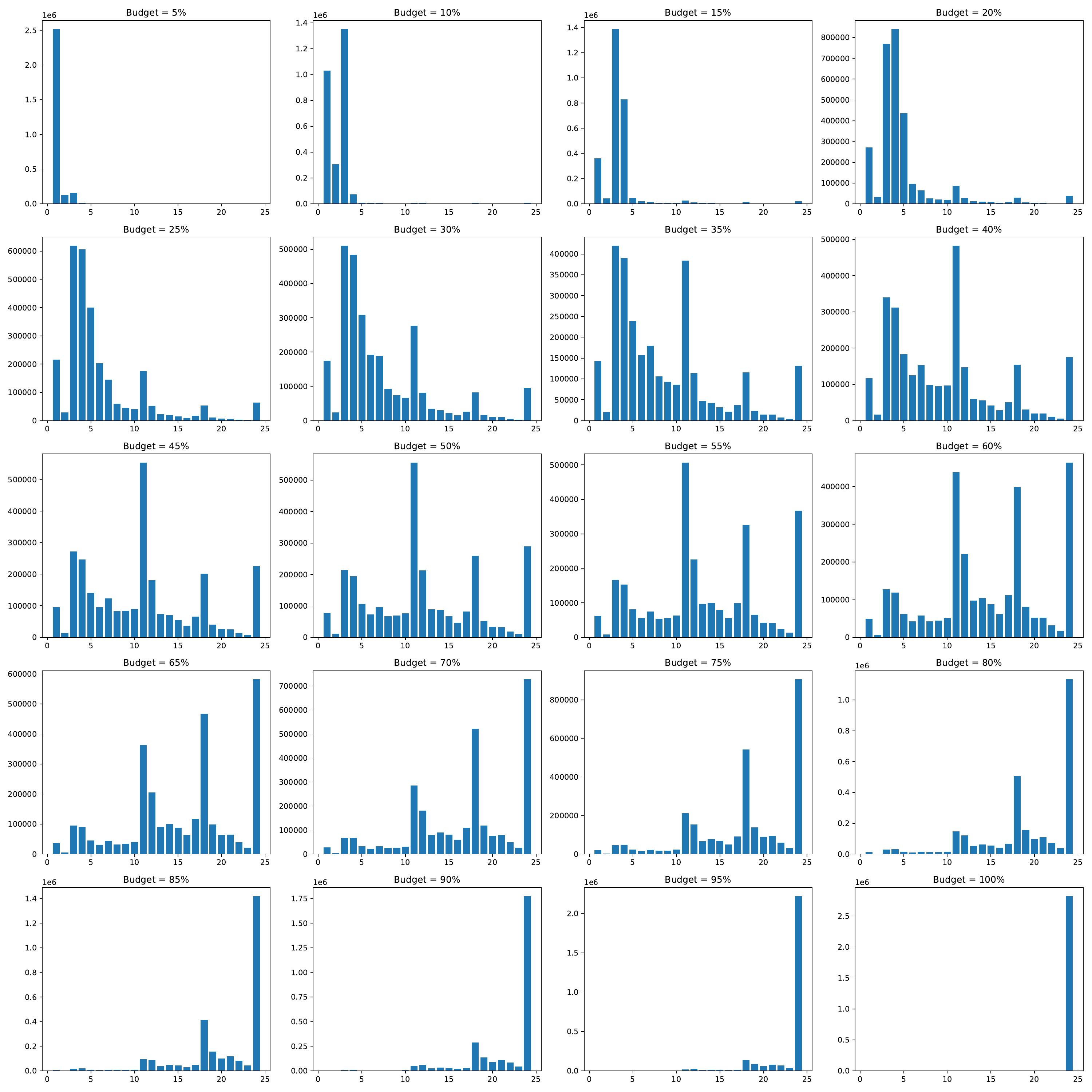}
    \caption{
    Distributions of the highest COMET layer scores calculated by UCB bandit (start layer = 1, $\gamma = 1.0$) across different budgets. 
    The distributions are plotted for $5\%$ budget increments, from $5\%$ to $100\%$, on the test set with the candidates generated via sampling. 
    }
    \label{fig:bandit_layers_viewed}
\end{figure*}

\section{Reranking Results in Tabular Format}
\label{app:reranking_results_table}

\Cref{tab:bandit} provides the results from \Cref{fig:bandit} in tabular format.

\begin{table*}
\newcolumntype{Y}{p{6mm}}
\small
\centering
\begin{minipage}{0.2\linewidth}
\centering
\vspace{1em}
Beam-Search\\
Final candidate is top-1
\end{minipage}
\begin{tabular}{lYYYYYYYYY}
\toprule
 & 15\% & 25\% & 35\% & 45\% & 55\% & 65\% & 75\% & 85\% & 95\%\\
\midrule
Bandit ($\gamma=1.0$) & 
28.4\% & 31.1\% & 45.1\% & 61.3\% & 78.2\% & 89.2\% & 95.3\% & 98.4\% & 99.7\%\\
Random & 
14.7\% & 23.6\% & 34.6\% & 44.7\% & 54.8\% & 64.3\% & 74.9\% & 85.3\% & 94.6\%\\
LogProb Sum & 
17.1\% & 27.2\% & 36.9\% & 46.8\% & 56.6\% & 66.2\% & 76.2\% & 86.0\% & 95.7\%\\
LogProb Avg & 
13.7\% & 23.2\% & 32.5\% & 42.1\% & 51.9\% & 62.3\% & 72.4\% & 83.2\% & 94.4\%\\
\bottomrule\end{tabular}

\begin{minipage}{0.2\linewidth}
\centering
Beam-Search\\
Final candidate score
\end{minipage}
\begin{tabular}{lYYYYYYYYY}
Bandit ($\gamma=1.0$) & 
78.74 & 79.12 & 79.59 & 79.91 & 80.11 & 80.18 & 80.20 & 80.21 & 80.21\\
Random & 
78.62 & 79.13 & 79.45 & 79.65 & 79.80 & 79.92 & 80.03 & 80.11 & 80.18\\
LogProb Sum & 
78.47 & 79.03 & 79.37 & 79.59 & 79.76 & 79.89 & 80.01 & 80.10 & 80.18\\
LogProb Avg & 
78.34 & 78.91 & 79.23 & 79.47 & 79.65 & 79.80 & 79.93 & 80.05 & 80.16\\
\bottomrule\end{tabular}

\begin{minipage}{0.2\linewidth}
\centering
Sampling\\
Final candidate is top-1
\end{minipage}
\begin{tabular}{lYYYYYYYYY}
Bandit ($\gamma=1.0$) & 
32.1\% & 33.9\% & 44.3\% & 57.1\% & 71.7\% & 83.7\% & 92.7\% & 97.5\% & 99.5\%\\
Random & 
15.1\% & 25.0\% & 35.8\% & 44.6\% & 55.9\% & 63.9\% & 75.4\% & 84.0\% & 95.2\%\\
LogProb Sum & 
15.8\% & 25.6\% & 34.3\% & 43.4\% & 52.6\% & 62.4\% & 72.4\% & 83.4\% & 93.4\%\\
LogProb Avg & 
9.3\% & 17.6\% & 27.1\% & 36.5\% & 47.1\% & 57.9\% & 68.8\% & 79.5\% & 92.1\%\\
\bottomrule\end{tabular}

\begin{minipage}{0.2\linewidth}
\centering
Sampling\\
Final candidate score
\end{minipage}
\begin{tabular}{lYYYYYYYYY}
Bandit ($\gamma=1.0$) & 
77.87 & 78.22 & 78.61 & 78.85 & 79.04 & 79.10 & 79.11 & 79.11 & 79.11\\
Random & 
77.99 & 78.33 & 78.58 & 78.70 & 78.84 & 78.90 & 78.98 & 79.03 & 79.08\\
LogProb Sum & 
77.59 & 78.08 & 78.36 & 78.54 & 78.70 & 78.83 & 78.92 & 79.00 & 79.07\\
LogProb Avg & 
77.41 & 77.88 & 78.16 & 78.38 & 78.54 & 78.70 & 78.82 & 78.94 & 79.04\\
\bottomrule\end{tabular}

\caption{
    Quality of the candidates returned by the Upper Confidence Bound bandit. Quality is measured in terms of the average final candidate score and the proportion to top-1 candidates selected. We plot these measures for various evaluation budgets. Cost (or budget) is given relative to calculating the full COMET scores for all candidates ($100\%$).
    Visualized in \Cref{fig:bandit}.
}
\label{tab:bandit}
\end{table*}

\begin{table*}[ht]
\newcolumntype{V}{>{\raggedleft\arraybackslash}p{6.8mm}}
\newcolumntype{Z}{>{\raggedleft\arraybackslash}p{8.3mm}}
\small
\setlength{\tabcolsep}{3pt}

\centering
\begin{tabular}{llVVVVVVVVVVVVVZ}
\parbox[t]{2mm}{\multirow{18}{*}{\rotatebox[origin=c]{90}{\bf Layer}}}
& \multicolumn{13}{c}{\bf \color{green!40!black} Layers} 
& \multicolumn{2}{r}{\bf \color{purple!80!black} Human} \\
 &  & \tiny 01\,\,\,\, & \tiny 03\,\,\,\, & \tiny 05\,\,\,\, & \tiny 07\,\,\,\, & \tiny 09\,\,\,\, & \tiny 11\,\,\,\, & \tiny 13\,\,\,\, & \tiny 15\,\,\,\, & \tiny 17\,\,\,\, & \tiny 19\,\,\,\, & \tiny 21\,\,\,\, & \tiny 23\,\,\,\, & \tiny 24\,\,\,\,\\
& \tiny 01
& \cellcolor{green!40} 1.00 & \cellcolor{green!16} 0.28 & \cellcolor{green!14} 0.23 & \cellcolor{green!16} 0.27 & \cellcolor{green!12} 0.17 & \cellcolor{green!11} 0.14 & \cellcolor{green!10} 0.10 & \cellcolor{green!8} 0.04 & \cellcolor{green!6} -0.01 & \cellcolor{green!5} -0.05 & \cellcolor{green!4} -0.07 & \cellcolor{green!4} -0.08 & \cellcolor{green!5} -0.06 & \cellcolor{purple!0} -0.033
\\
& \tiny 03
& \cellcolor{green!16} 0.28 & \cellcolor{green!40} 1.00 & \cellcolor{green!16} 0.28 & \cellcolor{green!10} 0.10 & \cellcolor{green!11} 0.12 & \cellcolor{green!9} 0.08 & \cellcolor{green!9} 0.06 & \cellcolor{green!9} 0.07 & \cellcolor{green!9} 0.08 & \cellcolor{green!10} 0.11 & \cellcolor{green!10} 0.10 & \cellcolor{green!8} 0.05 & \cellcolor{green!6} -0.01 & \cellcolor{purple!0} -0.014
\\
& \tiny 05
& \cellcolor{green!14} 0.23 & \cellcolor{green!16} 0.28 & \cellcolor{green!40} 1.00 & \cellcolor{green!37} 0.90 & \cellcolor{green!35} 0.86 & \cellcolor{green!33} 0.79 & \cellcolor{green!29} 0.68 & \cellcolor{green!26} 0.58 & \cellcolor{green!23} 0.48 & \cellcolor{green!21} 0.42 & \cellcolor{green!19} 0.38 & \cellcolor{green!17} 0.30 & \cellcolor{green!16} 0.28 & \cellcolor{purple!7} 0.064
\\
& \tiny 07
& \cellcolor{green!16} 0.27 & \cellcolor{green!10} 0.10 & \cellcolor{green!37} 0.90 & \cellcolor{green!40} 1.00 & \cellcolor{green!39} 0.97 & \cellcolor{green!37} 0.92 & \cellcolor{green!35} 0.84 & \cellcolor{green!31} 0.73 & \cellcolor{green!27} 0.61 & \cellcolor{green!24} 0.52 & \cellcolor{green!23} 0.48 & \cellcolor{green!19} 0.38 & \cellcolor{green!19} 0.36 & \cellcolor{purple!10} 0.096
\\
& \tiny 09
& \cellcolor{green!12} 0.17 & \cellcolor{green!11} 0.12 & \cellcolor{green!35} 0.86 & \cellcolor{green!39} 0.97 & \cellcolor{green!40} 1.00 & \cellcolor{green!39} 0.98 & \cellcolor{green!37} 0.90 & \cellcolor{green!33} 0.80 & \cellcolor{green!29} 0.68 & \cellcolor{green!27} 0.61 & \cellcolor{green!25} 0.56 & \cellcolor{green!22} 0.45 & \cellcolor{green!20} 0.41 & \cellcolor{purple!11} 0.116
\\
& \tiny 11
& \cellcolor{green!11} 0.14 & \cellcolor{green!9} 0.08 & \cellcolor{green!33} 0.79 & \cellcolor{green!37} 0.92 & \cellcolor{green!39} 0.98 & \cellcolor{green!40} 1.00 & \cellcolor{green!39} 0.96 & \cellcolor{green!35} 0.86 & \cellcolor{green!32} 0.75 & \cellcolor{green!29} 0.68 & \cellcolor{green!28} 0.63 & \cellcolor{green!24} 0.51 & \cellcolor{green!22} 0.46 & \cellcolor{purple!13} 0.132
\\
& \tiny 13
& \cellcolor{green!10} 0.10 & \cellcolor{green!9} 0.06 & \cellcolor{green!29} 0.68 & \cellcolor{green!35} 0.84 & \cellcolor{green!37} 0.90 & \cellcolor{green!39} 0.96 & \cellcolor{green!40} 1.00 & \cellcolor{green!38} 0.95 & \cellcolor{green!35} 0.86 & \cellcolor{green!33} 0.78 & \cellcolor{green!31} 0.73 & \cellcolor{green!27} 0.60 & \cellcolor{green!24} 0.53 & \cellcolor{purple!16} 0.176
\\
& \tiny 15
& \cellcolor{green!8} 0.04 & \cellcolor{green!9} 0.07 & \cellcolor{green!26} 0.58 & \cellcolor{green!31} 0.73 & \cellcolor{green!33} 0.80 & \cellcolor{green!35} 0.86 & \cellcolor{green!38} 0.95 & \cellcolor{green!40} 1.00 & \cellcolor{green!39} 0.97 & \cellcolor{green!37} 0.92 & \cellcolor{green!36} 0.88 & \cellcolor{green!31} 0.74 & \cellcolor{green!28} 0.65 & \cellcolor{purple!21} 0.230
\\
& \tiny 17
& \cellcolor{green!6} -0.01 & \cellcolor{green!9} 0.08 & \cellcolor{green!23} 0.48 & \cellcolor{green!27} 0.61 & \cellcolor{green!29} 0.68 & \cellcolor{green!32} 0.75 & \cellcolor{green!35} 0.86 & \cellcolor{green!39} 0.97 & \cellcolor{green!40} 1.00 & \cellcolor{green!39} 0.98 & \cellcolor{green!39} 0.96 & \cellcolor{green!34} 0.83 & \cellcolor{green!31} 0.73 & \cellcolor{purple!24} 0.264
\\
& \tiny 19
& \cellcolor{green!5} -0.05 & \cellcolor{green!10} 0.11 & \cellcolor{green!21} 0.42 & \cellcolor{green!24} 0.52 & \cellcolor{green!27} 0.61 & \cellcolor{green!29} 0.68 & \cellcolor{green!33} 0.78 & \cellcolor{green!37} 0.92 & \cellcolor{green!39} 0.98 & \cellcolor{green!40} 1.00 & \cellcolor{green!40} 0.99 & \cellcolor{green!35} 0.86 & \cellcolor{green!32} 0.75 & \cellcolor{purple!25} 0.273
\\
& \tiny 21
& \cellcolor{green!4} -0.07 & \cellcolor{green!10} 0.10 & \cellcolor{green!19} 0.38 & \cellcolor{green!23} 0.48 & \cellcolor{green!25} 0.56 & \cellcolor{green!28} 0.63 & \cellcolor{green!31} 0.73 & \cellcolor{green!36} 0.88 & \cellcolor{green!39} 0.96 & \cellcolor{green!40} 0.99 & \cellcolor{green!40} 1.00 & \cellcolor{green!37} 0.91 & \cellcolor{green!33} 0.80 & \cellcolor{purple!26} 0.283
\\
& \tiny 23
& \cellcolor{green!4} -0.08 & \cellcolor{green!8} 0.05 & \cellcolor{green!17} 0.30 & \cellcolor{green!19} 0.38 & \cellcolor{green!22} 0.45 & \cellcolor{green!24} 0.51 & \cellcolor{green!27} 0.60 & \cellcolor{green!31} 0.74 & \cellcolor{green!34} 0.83 & \cellcolor{green!35} 0.86 & \cellcolor{green!37} 0.91 & \cellcolor{green!40} 1.00 & \cellcolor{green!39} 0.96 & \cellcolor{purple!29} 0.319
\\
& \tiny 24
& \cellcolor{green!5} -0.06 & \cellcolor{green!6} -0.01 & \cellcolor{green!16} 0.28 & \cellcolor{green!19} 0.36 & \cellcolor{green!20} 0.41 & \cellcolor{green!22} 0.46 & \cellcolor{green!24} 0.53 & \cellcolor{green!28} 0.65 & \cellcolor{green!31} 0.73 & \cellcolor{green!32} 0.75 & \cellcolor{green!33} 0.80 & \cellcolor{green!39} 0.96 & \cellcolor{green!40} 1.00 & \cellcolor{purple!29} 0.327
\\

\end{tabular} \\
Baseline COMET
\vspace{10mm}

\begin{tabular}{llVVVVVVVVVVVVVZ}
\parbox[t]{2mm}{\multirow{18}{*}{\rotatebox[origin=c]{90}{\bf Layer}}}
& \multicolumn{13}{c}{\bf \color{green!40!black} Layers} 
& \multicolumn{2}{r}{\bf \color{purple!80!black} Human} \\
 &  & \tiny 01\,\,\,\, & \tiny 03\,\,\,\, & \tiny 05\,\,\,\, & \tiny 07\,\,\,\, & \tiny 09\,\,\,\, & \tiny 11\,\,\,\, & \tiny 13\,\,\,\, & \tiny 15\,\,\,\, & \tiny 17\,\,\,\, & \tiny 19\,\,\,\, & \tiny 21\,\,\,\, & \tiny 23\,\,\,\, & \tiny 24\,\,\,\,\\
& \tiny 01
& \cellcolor{green!40} 1.00 & \cellcolor{green!19} 0.37 & \cellcolor{green!17} 0.30 & \cellcolor{green!15} 0.24 & \cellcolor{green!14} 0.23 & \cellcolor{green!14} 0.21 & \cellcolor{green!12} 0.17 & \cellcolor{green!12} 0.17 & \cellcolor{green!12} 0.17 & \cellcolor{green!12} 0.15 & \cellcolor{green!12} 0.15 & \cellcolor{green!12} 0.15 & \cellcolor{green!12} 0.15 & \cellcolor{purple!4} 0.034
\\
& \tiny 03
& \cellcolor{green!19} 0.37 & \cellcolor{green!40} 1.00 & \cellcolor{green!32} 0.77 & \cellcolor{green!29} 0.67 & \cellcolor{green!28} 0.65 & \cellcolor{green!28} 0.64 & \cellcolor{green!25} 0.54 & \cellcolor{green!24} 0.52 & \cellcolor{green!24} 0.52 & \cellcolor{green!23} 0.49 & \cellcolor{green!23} 0.49 & \cellcolor{green!23} 0.50 & \cellcolor{green!23} 0.49 & \cellcolor{purple!15} 0.159
\\
& \tiny 05
& \cellcolor{green!17} 0.30 & \cellcolor{green!32} 0.77 & \cellcolor{green!40} 1.00 & \cellcolor{green!38} 0.95 & \cellcolor{green!38} 0.93 & \cellcolor{green!34} 0.82 & \cellcolor{green!31} 0.72 & \cellcolor{green!30} 0.70 & \cellcolor{green!30} 0.70 & \cellcolor{green!29} 0.66 & \cellcolor{green!29} 0.66 & \cellcolor{green!29} 0.66 & \cellcolor{green!28} 0.65 & \cellcolor{purple!19} 0.207
\\
& \tiny 07
& \cellcolor{green!15} 0.24 & \cellcolor{green!29} 0.67 & \cellcolor{green!38} 0.95 & \cellcolor{green!40} 1.00 & \cellcolor{green!40} 0.99 & \cellcolor{green!36} 0.87 & \cellcolor{green!32} 0.77 & \cellcolor{green!32} 0.75 & \cellcolor{green!31} 0.74 & \cellcolor{green!30} 0.70 & \cellcolor{green!30} 0.69 & \cellcolor{green!30} 0.69 & \cellcolor{green!29} 0.68 & \cellcolor{purple!20} 0.221
\\
& \tiny 09
& \cellcolor{green!14} 0.23 & \cellcolor{green!28} 0.65 & \cellcolor{green!38} 0.93 & \cellcolor{green!40} 0.99 & \cellcolor{green!40} 1.00 & \cellcolor{green!36} 0.88 & \cellcolor{green!33} 0.78 & \cellcolor{green!32} 0.76 & \cellcolor{green!32} 0.75 & \cellcolor{green!30} 0.71 & \cellcolor{green!30} 0.70 & \cellcolor{green!30} 0.70 & \cellcolor{green!30} 0.69 & \cellcolor{purple!20} 0.221
\\
& \tiny 11
& \cellcolor{green!14} 0.21 & \cellcolor{green!28} 0.64 & \cellcolor{green!34} 0.82 & \cellcolor{green!36} 0.87 & \cellcolor{green!36} 0.88 & \cellcolor{green!40} 1.00 & \cellcolor{green!38} 0.94 & \cellcolor{green!37} 0.91 & \cellcolor{green!36} 0.89 & \cellcolor{green!34} 0.82 & \cellcolor{green!34} 0.81 & \cellcolor{green!34} 0.81 & \cellcolor{green!33} 0.80 & \cellcolor{purple!23} 0.251
\\
& \tiny 13
& \cellcolor{green!12} 0.17 & \cellcolor{green!25} 0.54 & \cellcolor{green!31} 0.72 & \cellcolor{green!32} 0.77 & \cellcolor{green!33} 0.78 & \cellcolor{green!38} 0.94 & \cellcolor{green!40} 1.00 & \cellcolor{green!39} 0.98 & \cellcolor{green!39} 0.97 & \cellcolor{green!36} 0.88 & \cellcolor{green!36} 0.87 & \cellcolor{green!35} 0.86 & \cellcolor{green!35} 0.85 & \cellcolor{purple!25} 0.278
\\
& \tiny 15
& \cellcolor{green!12} 0.17 & \cellcolor{green!24} 0.52 & \cellcolor{green!30} 0.70 & \cellcolor{green!32} 0.75 & \cellcolor{green!32} 0.76 & \cellcolor{green!37} 0.91 & \cellcolor{green!39} 0.98 & \cellcolor{green!40} 1.00 & \cellcolor{green!40} 0.99 & \cellcolor{green!37} 0.91 & \cellcolor{green!36} 0.89 & \cellcolor{green!36} 0.89 & \cellcolor{green!36} 0.88 & \cellcolor{purple!25} 0.281
\\
& \tiny 17
& \cellcolor{green!12} 0.17 & \cellcolor{green!24} 0.52 & \cellcolor{green!30} 0.70 & \cellcolor{green!31} 0.74 & \cellcolor{green!32} 0.75 & \cellcolor{green!36} 0.89 & \cellcolor{green!39} 0.97 & \cellcolor{green!40} 0.99 & \cellcolor{green!40} 1.00 & \cellcolor{green!37} 0.92 & \cellcolor{green!37} 0.91 & \cellcolor{green!37} 0.90 & \cellcolor{green!36} 0.89 & \cellcolor{purple!26} 0.281
\\
& \tiny 19
& \cellcolor{green!12} 0.15 & \cellcolor{green!23} 0.49 & \cellcolor{green!29} 0.66 & \cellcolor{green!30} 0.70 & \cellcolor{green!30} 0.71 & \cellcolor{green!34} 0.82 & \cellcolor{green!36} 0.88 & \cellcolor{green!37} 0.91 & \cellcolor{green!37} 0.92 & \cellcolor{green!40} 1.00 & \cellcolor{green!40} 0.99 & \cellcolor{green!39} 0.98 & \cellcolor{green!39} 0.98 & \cellcolor{purple!28} 0.310
\\
& \tiny 21
& \cellcolor{green!12} 0.15 & \cellcolor{green!23} 0.49 & \cellcolor{green!29} 0.66 & \cellcolor{green!30} 0.69 & \cellcolor{green!30} 0.70 & \cellcolor{green!34} 0.81 & \cellcolor{green!36} 0.87 & \cellcolor{green!36} 0.89 & \cellcolor{green!37} 0.91 & \cellcolor{green!40} 0.99 & \cellcolor{green!40} 1.00 & \cellcolor{green!40} 1.00 & \cellcolor{green!40} 0.99 & \cellcolor{purple!28} 0.312
\\
& \tiny 23
& \cellcolor{green!12} 0.15 & \cellcolor{green!23} 0.50 & \cellcolor{green!29} 0.66 & \cellcolor{green!30} 0.69 & \cellcolor{green!30} 0.70 & \cellcolor{green!34} 0.81 & \cellcolor{green!35} 0.86 & \cellcolor{green!36} 0.89 & \cellcolor{green!37} 0.90 & \cellcolor{green!39} 0.98 & \cellcolor{green!40} 1.00 & \cellcolor{green!40} 1.00 & \cellcolor{green!40} 1.00 & \cellcolor{purple!28} 0.310
\\
& \tiny 24
& \cellcolor{green!12} 0.15 & \cellcolor{green!23} 0.49 & \cellcolor{green!28} 0.65 & \cellcolor{green!29} 0.68 & \cellcolor{green!30} 0.69 & \cellcolor{green!33} 0.80 & \cellcolor{green!35} 0.85 & \cellcolor{green!36} 0.88 & \cellcolor{green!36} 0.89 & \cellcolor{green!39} 0.98 & \cellcolor{green!40} 0.99 & \cellcolor{green!40} 1.00 & \cellcolor{green!40} 1.00 & \cellcolor{purple!28} 0.309
\\

\end{tabular} \\
Early-Exit COMET
\vspace{10mm}

\begin{tabular}{llVVVVVVVVVVVVVZ}
\parbox[t]{2mm}{\multirow{18}{*}{\rotatebox[origin=c]{90}{\bf Layer}}}
& \multicolumn{13}{c}{\bf \color{green!40!black} Layers} 
& \multicolumn{2}{r}{\bf \color{purple!80!black} Human} \\
 &  & \tiny 01\,\,\,\, & \tiny 03\,\,\,\, & \tiny 05\,\,\,\, & \tiny 07\,\,\,\, & \tiny 09\,\,\,\, & \tiny 11\,\,\,\, & \tiny 13\,\,\,\, & \tiny 15\,\,\,\, & \tiny 17\,\,\,\, & \tiny 19\,\,\,\, & \tiny 21\,\,\,\, & \tiny 23\,\,\,\, & \tiny 24\,\,\,\,\\
& \tiny 01
& \cellcolor{green!40} 1.00 & \cellcolor{green!11} 0.14 & \cellcolor{green!1} -0.17 & \cellcolor{green!1} -0.17 & \cellcolor{green!2} -0.14 & \cellcolor{green!2} -0.14 & \cellcolor{green!2} -0.14 & \cellcolor{green!2} -0.14 & \cellcolor{green!2} -0.15 & \cellcolor{green!2} -0.13 & \cellcolor{green!3} -0.12 & \cellcolor{green!3} -0.12 & \cellcolor{green!3} -0.12 & \cellcolor{purple!0} -0.057
\\
& \tiny 03
& \cellcolor{green!11} 0.14 & \cellcolor{green!40} 1.00 & \cellcolor{green!10} 0.11 & \cellcolor{green!10} 0.10 & \cellcolor{green!11} 0.12 & \cellcolor{green!10} 0.09 & \cellcolor{green!8} 0.05 & \cellcolor{green!8} 0.05 & \cellcolor{green!8} 0.05 & \cellcolor{green!9} 0.06 & \cellcolor{green!8} 0.03 & \cellcolor{green!8} 0.03 & \cellcolor{green!8} 0.03 & \cellcolor{purple!3} 0.018
\\
& \tiny 05
& \cellcolor{green!1} -0.17 & \cellcolor{green!10} 0.11 & \cellcolor{green!40} 1.00 & \cellcolor{green!35} 0.86 & \cellcolor{green!33} 0.78 & \cellcolor{green!30} 0.71 & \cellcolor{green!29} 0.66 & \cellcolor{green!28} 0.65 & \cellcolor{green!28} 0.64 & \cellcolor{green!26} 0.58 & \cellcolor{green!25} 0.56 & \cellcolor{green!25} 0.56 & \cellcolor{green!25} 0.56 & \cellcolor{purple!17} 0.186
\\
& \tiny 07
& \cellcolor{green!1} -0.17 & \cellcolor{green!10} 0.10 & \cellcolor{green!35} 0.86 & \cellcolor{green!40} 1.00 & \cellcolor{green!39} 0.97 & \cellcolor{green!33} 0.80 & \cellcolor{green!31} 0.74 & \cellcolor{green!31} 0.73 & \cellcolor{green!31} 0.73 & \cellcolor{green!29} 0.66 & \cellcolor{green!28} 0.63 & \cellcolor{green!28} 0.63 & \cellcolor{green!27} 0.62 & \cellcolor{purple!20} 0.220
\\
& \tiny 09
& \cellcolor{green!2} -0.14 & \cellcolor{green!11} 0.12 & \cellcolor{green!33} 0.78 & \cellcolor{green!39} 0.97 & \cellcolor{green!40} 1.00 & \cellcolor{green!35} 0.84 & \cellcolor{green!32} 0.77 & \cellcolor{green!32} 0.76 & \cellcolor{green!32} 0.76 & \cellcolor{green!29} 0.68 & \cellcolor{green!28} 0.65 & \cellcolor{green!28} 0.65 & \cellcolor{green!28} 0.65 & \cellcolor{purple!21} 0.231
\\
& \tiny 11
& \cellcolor{green!2} -0.14 & \cellcolor{green!10} 0.09 & \cellcolor{green!30} 0.71 & \cellcolor{green!33} 0.80 & \cellcolor{green!35} 0.84 & \cellcolor{green!40} 1.00 & \cellcolor{green!38} 0.95 & \cellcolor{green!37} 0.92 & \cellcolor{green!37} 0.91 & \cellcolor{green!34} 0.81 & \cellcolor{green!33} 0.78 & \cellcolor{green!32} 0.77 & \cellcolor{green!32} 0.77 & \cellcolor{purple!24} 0.267
\\
& \tiny 13
& \cellcolor{green!2} -0.14 & \cellcolor{green!8} 0.05 & \cellcolor{green!29} 0.66 & \cellcolor{green!31} 0.74 & \cellcolor{green!32} 0.77 & \cellcolor{green!38} 0.95 & \cellcolor{green!40} 1.00 & \cellcolor{green!40} 0.99 & \cellcolor{green!39} 0.98 & \cellcolor{green!36} 0.87 & \cellcolor{green!34} 0.83 & \cellcolor{green!34} 0.82 & \cellcolor{green!34} 0.82 & \cellcolor{purple!26} 0.284
\\
& \tiny 15
& \cellcolor{green!2} -0.14 & \cellcolor{green!8} 0.05 & \cellcolor{green!28} 0.65 & \cellcolor{green!31} 0.73 & \cellcolor{green!32} 0.76 & \cellcolor{green!37} 0.92 & \cellcolor{green!40} 0.99 & \cellcolor{green!40} 1.00 & \cellcolor{green!40} 0.99 & \cellcolor{green!36} 0.88 & \cellcolor{green!35} 0.84 & \cellcolor{green!34} 0.83 & \cellcolor{green!34} 0.83 & \cellcolor{purple!26} 0.285
\\
& \tiny 17
& \cellcolor{green!2} -0.15 & \cellcolor{green!8} 0.05 & \cellcolor{green!28} 0.64 & \cellcolor{green!31} 0.73 & \cellcolor{green!32} 0.76 & \cellcolor{green!37} 0.91 & \cellcolor{green!39} 0.98 & \cellcolor{green!40} 0.99 & \cellcolor{green!40} 1.00 & \cellcolor{green!36} 0.89 & \cellcolor{green!35} 0.85 & \cellcolor{green!35} 0.84 & \cellcolor{green!35} 0.84 & \cellcolor{purple!26} 0.287
\\
& \tiny 19
& \cellcolor{green!2} -0.13 & \cellcolor{green!9} 0.06 & \cellcolor{green!26} 0.58 & \cellcolor{green!29} 0.66 & \cellcolor{green!29} 0.68 & \cellcolor{green!34} 0.81 & \cellcolor{green!36} 0.87 & \cellcolor{green!36} 0.88 & \cellcolor{green!36} 0.89 & \cellcolor{green!40} 1.00 & \cellcolor{green!39} 0.97 & \cellcolor{green!39} 0.96 & \cellcolor{green!39} 0.96 & \cellcolor{purple!29} 0.320
\\
& \tiny 21
& \cellcolor{green!3} -0.12 & \cellcolor{green!8} 0.03 & \cellcolor{green!25} 0.56 & \cellcolor{green!28} 0.63 & \cellcolor{green!28} 0.65 & \cellcolor{green!33} 0.78 & \cellcolor{green!34} 0.83 & \cellcolor{green!35} 0.84 & \cellcolor{green!35} 0.85 & \cellcolor{green!39} 0.97 & \cellcolor{green!40} 1.00 & \cellcolor{green!40} 1.00 & \cellcolor{green!40} 0.99 & \cellcolor{purple!29} 0.324
\\
& \tiny 23
& \cellcolor{green!3} -0.12 & \cellcolor{green!8} 0.03 & \cellcolor{green!25} 0.56 & \cellcolor{green!28} 0.63 & \cellcolor{green!28} 0.65 & \cellcolor{green!32} 0.77 & \cellcolor{green!34} 0.82 & \cellcolor{green!34} 0.83 & \cellcolor{green!35} 0.84 & \cellcolor{green!39} 0.96 & \cellcolor{green!40} 1.00 & \cellcolor{green!40} 1.00 & \cellcolor{green!40} 1.00 & \cellcolor{purple!29} 0.324
\\
& \tiny 24
& \cellcolor{green!3} -0.12 & \cellcolor{green!8} 0.03 & \cellcolor{green!25} 0.56 & \cellcolor{green!27} 0.62 & \cellcolor{green!28} 0.65 & \cellcolor{green!32} 0.77 & \cellcolor{green!34} 0.82 & \cellcolor{green!34} 0.83 & \cellcolor{green!35} 0.84 & \cellcolor{green!39} 0.96 & \cellcolor{green!40} 0.99 & \cellcolor{green!40} 1.00 & \cellcolor{green!40} 1.00 & \cellcolor{purple!29} 0.325
\\

\end{tabular}\\
Early-Exit COMET (separate heads)

\caption{Pearson correlations between intermediate layer outputs (green) and between intermediate layer outputs and humans (purple) for unsupervised Early-Exit based on standard COMET, supervised Early-Exit with a single regression head, and supervised Early-Exit with separate regression head for each layer.}
\label{10-eval_all_big}
\end{table*}

\clearpage
\clearpage

\section{Partial COMET}
\label{sec:goal_partial}

Most quality estimation models, such as COMET, struggle with partial generations and other non-standard inputs \citep{zouhar-etal-2024-pitfalls}.
This is also crucial in applications like speech translation, where the input is often segmented into smaller chunks \citep{sperber-etal-2024-evaluating}, potentially resulting in partial sentences or incomplete translations that current quality estimation models struggle to handle effectively \citep[][Appendix D1]{amrhein-haddow-2022-dont}.

In this section, we introduce \textit{Partial COMET} (addressing non-standard input obstacle), ensuring reliable quality assessments for incomplete outputs. We show that also beyond evaluation, partial quality estimation can assess translation quality early in the translation generation process, and help discard unpromising candidates from beam search or sampling-based methods, allowing only the most promising candidates to be further expanded.
(\Cref{sec:goal_partial})

Machine translation quality estimation systems provide an assessment of model output quality.
However, in many cases, we wish to know the quality estimation even before a full translation is produced, for example in setups with unclear segmentations, such as simultaneous speech translation, or in the middle of the generation process.
Using a quality estimation system trained on full translation candidates on partial generations will lead to lower scores because the translation is technically not correct.
In this section, we propose a model that is able to score even partial generations, which are prefixes of the original translations.
This can be then used during beam search or batch sampling to discard unpromising candidates.

\subsection{Modeling}

For partial generations, we do not yet know the full translation, but only its prefix: $t_{<i}$.
Using this as an input to a quality estimation model that expects a full translation would result in a unfairly low scores.
Therefore, we explicitly train a function for partial generations $f_p$, which sees partial generations during training:
\begin{align}
\mathcal{L}_2(y, f(s, t_{<p}))
\end{align}
This way, the quality estimation model predicts final translation scores based on just the translation prefix.
See \Cref{ex:partial_generation}.

\begin{table}[ht]
\centering
\small
\begin{tabular}{lll}
\toprule
\parbox[t]{2mm}{\multirow{2}{*}{\rotatebox[origin=c]{90}{Orig.}}}
& Hi there! Let's go eat, no? & \multirow{2}{*}{$\rightarrow$ 65}\\
& Hallo du! Lass uns essen gehen, oder? & \\[0.5em]
\parbox[t]{2mm}{\multirow{2}{*}{\rotatebox[origin=c]{90}{Part.}}}
& Hi there! Let's go eat, no? & \multirow{2}{*}{$\rightarrow$ 65}\\
& Hallo du! Lass--- \\
\bottomrule
\end{tabular}
\captionof{example}{Setup for partial translation quality estimation. The quality estimation model sees only half the translation but predicts the original human score for the whole translation.}
\label{ex:partial_generation}
\end{table}

To make this applicable for generation in MT models, we need to know when to stop in the middle of the translation.
For this, we use a heuristic based on 25\%, 50\%, and 75\% of the source segment length.
To account for different language verbosities, we multiply the portion of the source length with fertility for each language pair, such as 1.1 for English$\rightarrow$German.
So for 50\%, the model sees $t_{<1.1\times |s| \times 50\%}$.
For each training segment we randomly choose whether 25\%, 50\%, 75\%, or 100\% is revealed.

\subsection{Results}

\paragraph{Full COMET falls short.}
We show the results, as measured by Pearson correlation across WMT23 in \Cref{tab:results_partial}.
The full COMET strongly underperforms on partial generations.
Therefore, this quality estimation model is not suitable for evaluating incomplete segments during generation.
In contrast, the partial quality estimation model correlates much more.
This can also be due to human annotators overfocusing on the beginning of the translations when providing assessment scores \citep{magooda2020attendbeginningstudyusing}, or the quality estimation system picking up spurious correlations that give away the score \citep{zouhar-etal-2024-pitfalls}.

\begin{table}[ht]
\small
\centering
\begin{tabular}{lcc}
\toprule
\bf Segments & \bf Full COMET & \bf Partial COMET \\
\midrule
Partial 25\% & 0.097 & 0.210 \\
Partial 50\% & 0.108 & 0.250 \\
Partial 75\% & 0.155 & 0.283 \\
Original 100\% & 0.327 & 0.318 \\
\bottomrule
\end{tabular}

\caption{Pearson correlation in original and partial (half length) translation evaluation setups. The models are either trained on full or partial generations.}
\label{tab:results_partial}
\end{table}

\paragraph{Partial COMET prunes generations.}

To show that the partial COMET is useful also in practice, we consider a generative model, such as for machine translation.
In this case, the model produces multiple generations at the same time either using parallel sampling or beam-search.
Higher beam count or parallel samples generally lead to better performance, but also take much more compute.
In this setup, partial COMET is useful because it can prune unpromising generations.

We use the the \href{https://huggingface.co/facebook/nllb-200-distilled-600M}{600M-parameter distilled NLLB model} \citep{nllb2022} with beam search and parallel sampling of 200\footnote{For computational tractability, we  run this evaluation on a subset of WMT 2023 data, sampling 2000 examples for each language pair.}.
However, instead of generating all 200 candidates until the end, we perform reranking with partial COMET when only 25\% of the candidate's output has been generated.
Then, we take only a fraction to continue in the generation of the main model.
At 50\%, we again perform the intermediate reranking, and once again at 75\%.

\begin{figure*}[ht]
    \centering
    Beam-Search\\
    \includegraphics[width=0.49\linewidth]{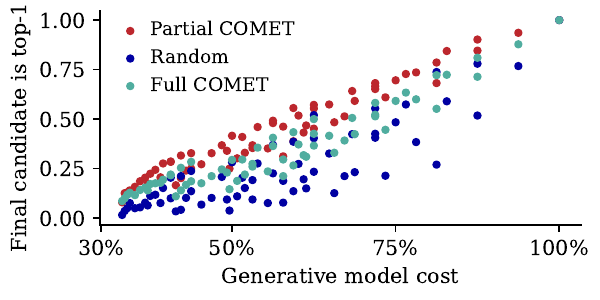}
    \includegraphics[width=0.49\linewidth]{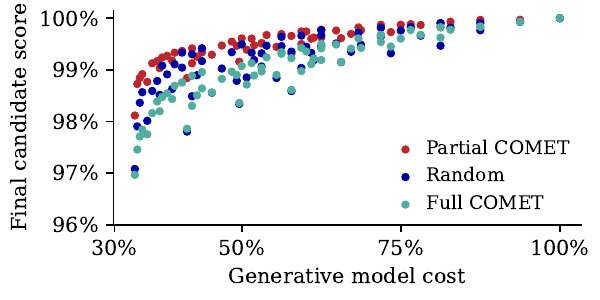}
    Sampling\\
    \includegraphics[width=0.49\linewidth]{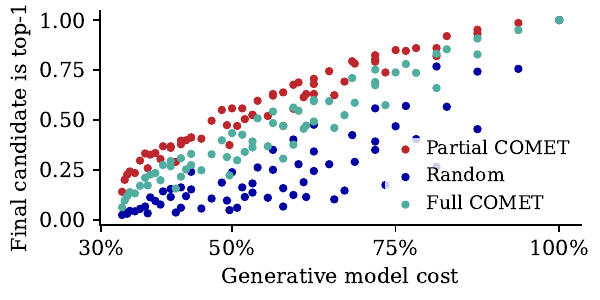}
    \includegraphics[width=0.49\linewidth]{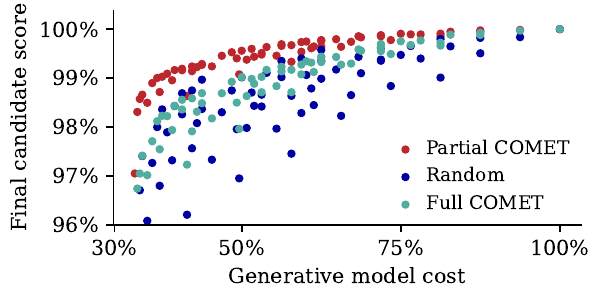}

    \caption{
    Proportion of the pruning process leading to the top candidate being chosen (left) or final candidate score (right) with respect to the computation cost of the generative model.
    }
    \label{fig:31-partial_candidates_top1_beam}
\end{figure*}

We try all combinations of pruning bottom 0\%, 25\%, 50\%, or 75\% at each of 25\%, 50\%, and 75\% target lengths.
The results in \Cref{fig:31-partial_candidates_top1_beam} show that by using Partial COMET, we are more likely to prune lower-quality candidates than with the original COMET, which itself performs better than random pruning.

\paragraph{Limitations.}
For Partial COMET, in some cases, the human score is not aligned with the translation substring, such as when the error which cause a lower human score is not present in the substring.
This can be further remedied by word-level quality estimation, where considering only a substring to evaluation also automatically select only the present word-level errors.

\subsection{Works on Segmentation-Robust QE}
Learned automated metrics, such as COMET or MetricX \citep{juraska-etal-2024-metricx} inherit all problems of statistical machine learning, by expecting the input to come from a particular distribution \citep{zouhar-etal-2024-pitfalls}.
\citet{amrhein-haddow-2022-dont} hint, and we confirm, that using a quality estimation model trained on full translations on incomplete segments leads to poor correlations.
Justifiably so, because from the perspective of the quality estimation model, the translation is thus incorrect.
This is problematic during decoding or in cases where the translation segmentation is unclear, such as for speech translation \citep{akhbardeh-etal-2021-findings-FIXED,salesky-etal-2023-evaluating,han-etal-2024-speechqe,sperber-etal-2024-evaluating}.

\end{document}